\pdfoutput=1
\documentclass[11pt]{article}

\usepackage[]{acl}

\usepackage{times}
\usepackage{latexsym}

\usepackage{verbatim}
\usepackage{amsmath}
\usepackage{multirow}
\usepackage{booktabs}
\usepackage{tabu}
\usepackage{pdfpages}
\usepackage{subfigure}
\usepackage{amsfonts,amssymb}
\usepackage{graphicx}
\usepackage[ruled,vlined]{algorithm2e}
\usepackage[T1]{fontenc}
\usepackage[utf8]{inputenc}

\usepackage{microtype}

\usepackage{inconsolata}

\title{ChatKBQA: A Generate-then-Retrieve Framework for Knowledge Base Question Answering with Fine-tuned Large Language Models}

\author{{\bf Haoran Luo\textsuperscript{\rm 1}, Haihong E\textsuperscript{\rm 1}\thanks{\ \ Corresponding author.}  , Zichen Tang\textsuperscript{\rm 1}, Shiyao Peng\textsuperscript{\rm 1}, Yikai Guo\textsuperscript{\rm 3}, Wentai Zhang\textsuperscript{\rm 1},} \\
{\bf Chenghao Ma\textsuperscript{\rm 1}, Guanting Dong\textsuperscript{\rm 2}, Meina Song\textsuperscript{\rm 1}, Wei Lin\textsuperscript{\rm 4}, Yifan Zhu\textsuperscript{\rm 1}, Luu Anh Tuan\textsuperscript{\rm 5}} \\
         \textsuperscript{1}School of Computer Science, Beijing University of Posts and Telecommunications, China \\ 
         \textsuperscript{2}School of Artificial Intelligence, Beijing University of Posts and Telecommunications, China \\ 
         \textsuperscript{3}Beijing Institute of Computer Technology and Application 
         \ \textsuperscript{4}Inspur Group Co., Ltd., China \\ 
         \textsuperscript{5}College of Computing and Data Science, Nanyang Technological University, Singapore \\ 
         \texttt{\{luohaoran, ehaihong, yifan\_zhu\}@bupt.edu.cn, anhtuan.luu@ntu.edu.sg}}

\begin{document}
\maketitle

\begin{abstract}
Knowledge Base Question Answering (KBQA) aims to answer natural language questions over large-scale knowledge bases (KBs), which can be summarized into two crucial steps: knowledge retrieval and semantic parsing. However, three core challenges remain: inefficient knowledge retrieval,  mistakes of retrieval adversely impacting semantic parsing, and the complexity of previous KBQA methods. To tackle these challenges, we introduce ChatKBQA, a novel and simple generate-then-retrieve KBQA framework, which proposes first generating the logical form with fine-tuned LLMs, then retrieving and replacing entities and relations with an unsupervised retrieval method, to improve both generation and retrieval more directly. Experimental results show that ChatKBQA achieves new state-of-the-art performance on standard KBQA datasets, WebQSP, and CWQ. This work can also be regarded as a new paradigm for combining LLMs with knowledge graphs (KGs) for interpretable and knowledge-required question answering. 
% Our code is publicly available at \url{https://anonymous.4open.science/r/ChatKBQA}.
Our code is publicly available\footnote{\url{https://github.com/LHRLAB/ChatKBQA}}.
\end{abstract}

\section{Introduction}

\begin{figure}[!t]
\centering
\includegraphics[width=7.8cm]{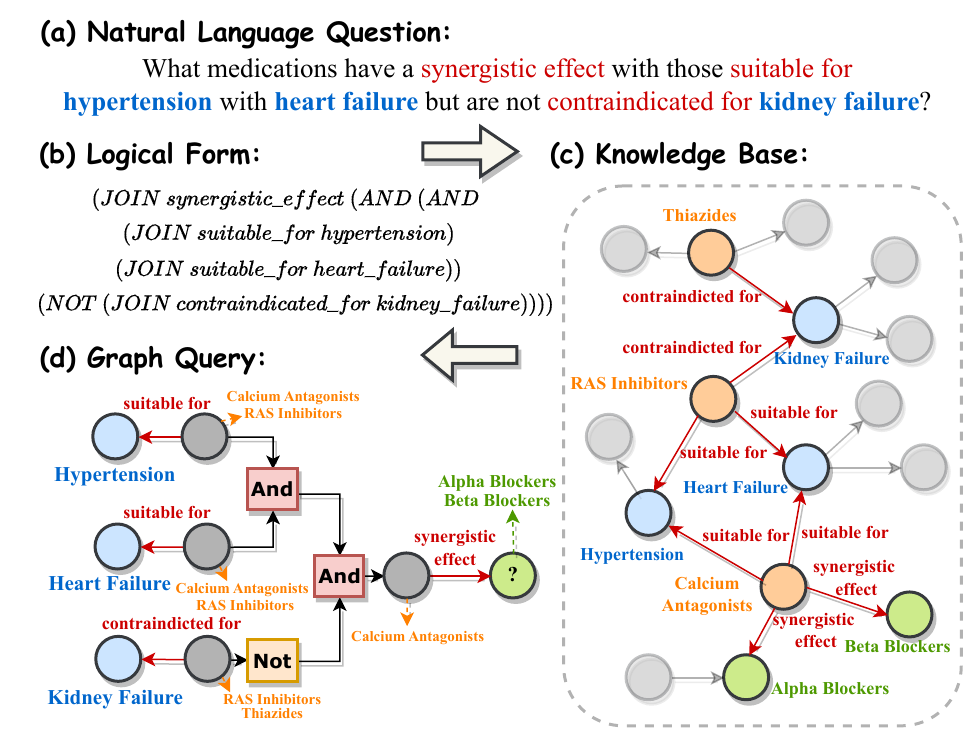}
\caption{An example of KBQA task to answer a natural language question by converting the question to a graph query which can be executed over Knowledge Base. }
\label{f0}
% \vspace{-2mm}
\end{figure}

Knowledge Base Question Answering (KBQA) is a classical NLP task to answer natural language questions based on facts over a large-scale knowledge base (KB), such as Freebase~\citep{Freebase}, Wikidata~\citep{Wikidata}, and DBpedia~\citep{DBpedia}, which are composed of structured knowledge graphs (KGs) built from triples consisting of (head entity, relation, tail entity). Previous KBQA methods primarily addressed two core issues: knowledge retrieval~\citep{KR} and semantic parsing~\citep{SP}. Knowledge retrieval mainly aims to locate the most relevant entities, relations, or triples according to the question from KB, to narrow the scope of consideration. Then, semantic parsing essentially converts the question from unstructured natural language into a structured logical form (such as S-expression~\citep{GrailQA}), which can then be converted into an executable graph database query (such as SPARQL~\citep{SPARQL}) to obtain precise answers and interpretable paths, as shown in Figure~\ref{f0}.

\begin{figure*}[!t]
\centering
\includegraphics[width=15.9cm]{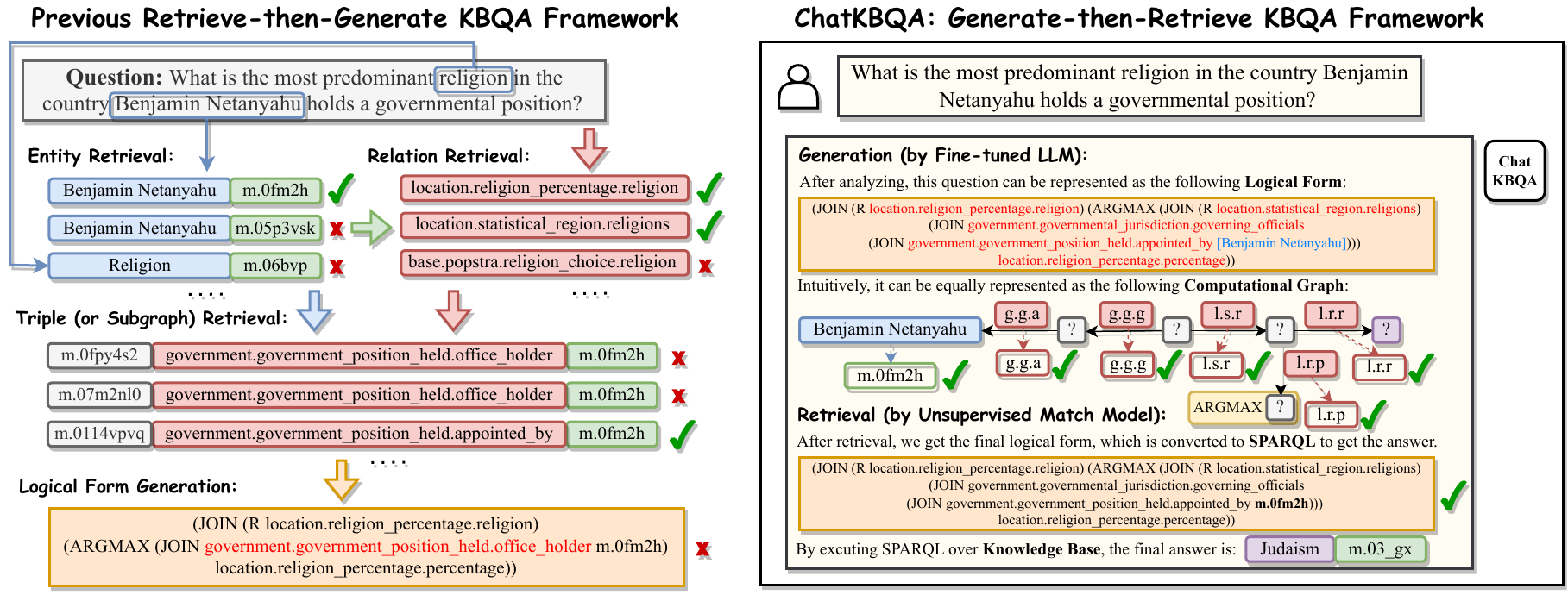}
\caption{Comparison of the previous retrieve-then-generate KBQA framework (left) and our proposed generate-then-retrieve KBQA framework, ChatKBQA (right). Note that labels such as "\texttt{g.g.a}" etc. in the computational graph are acronyms for relation names such as "\texttt{government.government\_position\_held.appointed\_by}". }
\label{f1}
\vspace{0.5mm}
\end{figure*}

Previous KBQA work~\citep{KV-Mem,PullNet,SR} proposed different knowledge retrieval methods with technologies of named entity recognition (NER)~\citep{BERT}, entity linking~\citep{ELQ} or subgraph retrieval~\citep{SR} to align natural language questions with structured KB. After retrieving factual triples, some studies~\citep{STAGG,QGG,UniKGQA} utilized strategies of step-wise query graph generation and search answers with semantic parsing. On the other hand, other work~\citep{RnG-KBQA,GMT-KBQA,TIARA,DECAF,FC-KBQA} performed semantic parsing by using a seq2seq model like T5~\citep{T5} to generate a logical form and then converted it to an SPARQL query to fetch answers when executed over KB.

Despite this, three main challenges remain, as shown on the left side of Figure~\ref{f1}: (i) \textbf{Low retrieval efficiency.} Traditional methods first identify the span of candidate entities and then do entity retrieval and relation retrieval. Since the structure of natural language questions differs from KB facts, most approaches require training dedicated models for extraction and linking inefficiently. (ii) \textbf{Incorrect retrieval results will mislead semantic parsing.} Previous methods have utilized retrieved triples also as input of reference to the seq2seq model along with the original question. However, since the retrieved triples are not always accurate, they adversely impact semantic parsing outcomes. Additionally, if there are numerous retrieved triples, the seq2seq model requires a much longer context length. (iii) \textbf{Multiple processing steps make KBQA a redundantly complex task.} Previous work decomposed the KBQA task into multiple sub-tasks~\citep{GMT-KBQA,TIARA,DECAF}, forming a complex pipeline, which made reproduction and migration challenging. In the era when large language models (LLMs)~\citep{GPT4,Llama2,LLMKGsurvey} are restructuring traditional NLP tasks~\citep{Flan-T5,LLMKGsurvey}, a more straightforward solution utilizing LLMs to reformulate the traditional KBQA paradigm is promising.

To overcome these challenges, we introduce \textbf{ChatKBQA}, a novel generate-then-retrieve KBQA framework based on open-source LLMs, such as Llama~\citep{Llama2}, ChatGLM~\citep{GLM} and Baichuan~\citep{Baichuan2}. As illustrated on the right side of Figure~\ref{f1}, ChatKBQA simplifies KBQA into two efficient phases: generating logical forms and then retrieving relevant entities and relations. \textbf{In the generation phase}, leveraging instruction tuning~\citep{PEFT}, fine-tuned LLMs exhibit high accuracy in semantic parsing of natural language questions without retrieval. The generated logical forms are not only mostly correct in skeleton (entities and relations masked) but also semantically consistent or close to the ground truth in terms of entities and relations. \textbf{In the retrieval phase}, ChatKBQA proposes an unsupervised retrieval method that employs phrase-level semantic retrieval within knowledge bases to improve generation accuracy and retrieval efficiency further. Additionally, ChatKBQA features a plug-and-play characteristic, ensuring compatibility with various LLMs and retrieval models, making it a flexible solution for KBQA tasks.

To valid the performance of our proposed framework, we conduct experiments on two standard KBQA datasets, WebQSP~\citep{STAGG} and ComplexWebQuestions (CWQ)~\citep{CWQ}, with both settings of using and not using golden entities. The experimental results demonstrate that ChatKBQA achieves a new state-of-the-art performance in the KBQA task. We also set up additional experiments to validate that our generate-then-retrieve approach improves both generation and retrieval results efficiency. 
Finally, we also discuss how insights from this framework lead us to envision future combinations of LLMs and KGs for knowledgable and interpretable Q\&A.

% Our main contributions are summarized as follows:
% \begin{itemize}
% \item We propose a straightforward KBQA framework that uses fine-tuned open-source large models for the first time.
% \item Innovatively, we adopt a generate-then-retrieve approach to enhance generation outcomes and retrieval efficiency separately, ultimately boosting KBQA performance.
% \item Our framework has plug-and-play capabilities, allowing flexible replacement of LLMs and retrieval models to address the KBQA challenge.
% \item Our approach introduces a new paradigm for LLMs to conduct interpretable external knowledge-based Q\&A, offering a fresh perspective on merging LLMs and KGs.
% \end{itemize}

\section{Related Work}
\label{s2}
\subsection{Knowledge Base Question Answering}
% 郭艺凯1115：这一部分应当说明这两者的好坏优劣，要向SP这一部分倾斜，然后引出我们室SP
Existing Knowledge Base Question Answering (KBQA) methods can be broadly categorized into Information Retrieval-based (IR-based) and Semantic Parsing-based (SP-based) methods. Recently, there have been some KBQA methods based on large language models (LLM-based) as well.

\textbf{(a) IR-based KBQA methods}~\citep{KV-Mem,PullNet,EmbedKGQA,NSM,TransferNet,SR} primarily retrieve relevant factual triples or text from KBs based on natural language questions, forming a subgraph to determine answers.

\textbf{(b) SP-based KBQA methods} focus on translating questions into logical forms executable against KBs, such as SPARQL, query graph, and S-expression. Some SP-based approaches~\citep{STAGG,UHop,TopicUnits,TextRay,QGG,UniKGQA} utilize strategies of \textbf{step-wise} query graph generation and search for semantic parsing. Alternatively, other SP-based methods~\citep{CBR-KBQA,RnG-KBQA,ProgramTransfer,TIARA,GMT-KBQA,UnifiedSKG,DECAF,FC-KBQA} employ \textbf{seq2seq} models to generate S-expressions completely and offer various enhancements to the semantic parsing process. 

\textbf{(c) LLM-based KBQA methods}~\citep{StructGPT,PanGu,ToG} utilize the thinking capabilities of LLMs to find answers by retrieving from the graph in a step-wise manner.

In this paper, our proposed ChatKBQA is the first SP-based KBQA method using fine-tuned LLMs, which innovatively proposes a generate-then-retrieve approach to simplify KBQA method.

\subsection{Large Language Models} 
% 郭艺凯1115：有点短，应该至少阐述1个观点，可以是从模型参数上，比如微调小模型，可以在特定任务产生超越大模型的效果
With ChatGPT~\citep{GPT4} displaying the prowess of decoder-only large language models (LLMs), many traditional NLP tasks are becoming simplified~\citep{LLMKGsurvey}. Subsequently, open-source LLMs like Llama~\citep{Llama2}, ChatGLM~\citep{GLM}, and Baichuan~\citep{Baichuan2} emerged and can be supervisedly fine-tuned (SFT) using Parameter-Efficient Fine-Tuning (PEFT) technologies~\citep{PEFT} such as LoRA~\citep{LoRA}, QLoRA~\citep{QLoRA}, P-Tuning v2~\citep{P-Tuningv2}, and Freeze~\citep{Freeze}, enhancing the capabilities of LLMs for specific tasks. 
% While LLMs often produce hallucinations~\citep{hallucinations}. Technologies like Chain-of-Thought (CoT)~\citep{CoT}, Tree of Thoughts (ToT)~\citep{ToT}, Graph-of-Thought (GoT)~\citep{GoT}, and Program of Thoughts (PoT)~\citep{PoT} have mitigated these hallucinations to some extent, but factual errors still occur~\citep{LLMKGsurvey}. Therefore, Graph Query of Thought (GQoT), generating explainable graph queries over KBs by fine-tuned LLMs, is a promising approach for interpretable and externally knowledge-required question answering. 
% In this paper, for the first time, ChatKBQA employs the instruction-tuning technique to fine-tune open-source LLMs, achieving impressive semantic parsing. It combines LLMs' powerful semantic parsing capabilities with KGs' interpretability advantages, introducing a new Graph Query of Thought (GQoT) paradigm for LLM+KG applications.

\subsection{Knowledge Retrieval for KBQA} General retrieval methods are typically divided into lexical methods, such as BM25~\citep{BM25}, and dense retrieval models, such as Dense Passage Retrieval (DPR)~\citep{DPR}, SimCSE~\citep{SimCSE}, and Contriever~\citep{Contriever}. In KBQA task, to better utilize knowledge related to the question from KB, ELQ~\citep{ELQ} and FACC1~\citep{FACC1} are commonly used to entity retrieval. 

In this paper, our ChatKBQA framework proposes a phrase-level retrieval method for entities and relations in an unsupervised manner after LLM's generation of logical form, improving both generation performance and retrieval efficiency. 
% Both CBR-KBQA and RnG-KBQA employ ELQ for dense-retrieval-based entity linking, while GMT-KBQA complements ELQ retrieval results with FACC1, enhancing the coverage of candidate entities. For relation retrieval, GMT-KBQA employs a two-stage relation retrieval module using a bi-encoder and cross-encoder architecture, and utilizes FAISS~\citep{FAISS} for neighboring relation retrieval. TIARA leverages a cross-encoder to learn interactive representations between the question and schema, ranking classes and relations based on matching scores.

\section{Preliminaries}
In this section, we define two basic concepts of our work: the knowledge base and the logical form, followed by the problem statement for KBQA task.
% \subsection{Problem Statement}

\textbf{Definition 1: Knowledge Base (KB).} A KB $\mathcal{K}=\{(s,r,o)|s \in \mathcal{E}, r \in \mathcal{R}, o \in \mathcal{E} \cup \mathcal{L}\}$ is an RDF graph consisting of triples $(s,r,o)$ where $s$ is an entity, $r$ is a relation , and $o$ can be an entity or a literal. Each entity $e \in \mathcal{E}$ in the entity set $\mathcal{E}$ is represented by a unique ID, e.g., \texttt{$e$.id="m.0fm2h"}, which can be queried to get its label as \texttt{$e$.label="Benjamin\ Netanyahu"}. Each relation $r \in \mathcal{R}$ in the relation set $\mathcal{R}$ has a multiple-level label, e.g. \texttt{$r$="government. government\_position\_held. appointed\_by"}. 
% Besides, a literal $l \in \mathcal{L}$ is usually "integer" (e.g., \texttt{$l$="32"}), "float" (e.g., \texttt{$l$="3.2"}), "year" (e.g., \texttt{$l$="1999"}), "year\&month" (e.g., \texttt{$l$="1999-12"}), or "date" (e.g., \texttt{$l$="1999-12-31"}).

\textbf{Definition 2: Logical Form. } A logical form is a structured representation of a natural language question. Taking the S-expression as an example, a logical form usually consists of projection and various operators. Projection operation represents a one-hop query of a triple $(s,r,o)$ on $s$ or $o$, where, $(?,r,o)$ is denoted as \texttt{(JOIN $r$ $o$)}, while $(s,r,?)$ is denoted as \texttt{(JOIN (R $r$) $s$)}. Other operators, e.g. "AND", "COUNT", and "ARGMAX", are introduced in Appendix~\ref{operator}.
% Various operators include "AND" \texttt{(AND $E_1$ $E_2$)} to denote taking the intersection of $E_1$ and $E_2$, "COUNT" \texttt{(COUNT $E_1$)} to denote counting $E_1$, "ARGMAX" \texttt{(ARGMAX $E_1$ $r$)} to denote taking the max literal obtained after the projection of $E_1$ in the $r$ relation, "ARGMIN" \texttt{(ARGMIN $E_1$ $r$) } to denote taking the min literal obtained after the projection of the $r$ relation for $E_1$, "GT" \texttt{(GT $E_1$ $l$)} means to take the portion of $E_1$ that is greater than $l$, "GE" \texttt {(GE $E_1$ $l$)} to denote taking the part of $E_1$ greater than or equal to $l$, "LT" \texttt{(LT $E_1$ $l$)} to denote taking the part of $E_1$ less than $l$, "LE" \texttt{(LE $E_1$ $l$)} to denote taking the part of $E_1$ which is less than or equal to $l$, where $E_1$ or $E_2$ denote a sublayer logical form.

\SetKwFunction{Convert}{Convert}
\SetKwFunction{Execute}{Execute}
\SetKwFunction{Sp}{Sp}
\textbf{Problem Statement. }For KBQA task, given a natural language question $Q$, and a knowledge base $\mathcal{K}$, we need to first convert $Q$ into a logical form $F=\Sp(Q)$, where $\Sp(.)$ is a semantic parsing function. Then convert $F$ to the equivalent SPARQL query $q=\Convert(F)$, where $\Convert(.)$ is the fixed conversion function. Finally, the final set of answers $A=\Execute(q|\mathcal{K})$ is obtained by executing $q$ against $\mathcal{K}$, where $\Execute(.)$ is the query execution function.

% \subsection{Efficient Fine-Tuning on LLMs}
% In order to reduce the cost of fine-tuning LLMs with a large number of parameters, Parameter Efficient Fine-Tuning (PEFT) methods have emerged to fine-tune only a small number of model parameters and achieve performance comparable to full fine-tuning. Among them, LoRA~\citep{LoRA} reduces the memory footprint of large language models with changing weights during fine-tuning by using a low-rank approximation.QLoRA~\citep{QLoRA} further reduces memory by back-propagating the gradient into a frozen 4-bit quantized model, while maintaining the performance of the full 16-bit fine-tuning task. P-tuning v2~\citep{P-Tuningv2} employs a prefix-tuning approach that incorporates fine-tunable parameters at each layer in front of the inputs.Freeze~\citep{Freeze} speeds up the model's convergence by fine-tuning only the fully-connected layer parameters of the last few layers of the Transformer while freezing all other parameters.
% \subsection{Unsupervised Retrieval Methods}
% Given a query, unsupervised retrieval methods require no additional training to select the top k semantically most similar of the candidate set as the set of retrieved answers. BM25~\citep{BM25} uses term frequencies and inverse document frequencies to rank documents based on their relevance to a given query. SimCSE~\citep{SimCSE} and Contriever~\citep{Contriever} are both unsupervised dense information retrieval using comparative learning models.
\begin{figure*}[h!t]
\centering
\includegraphics[width=15.8cm]{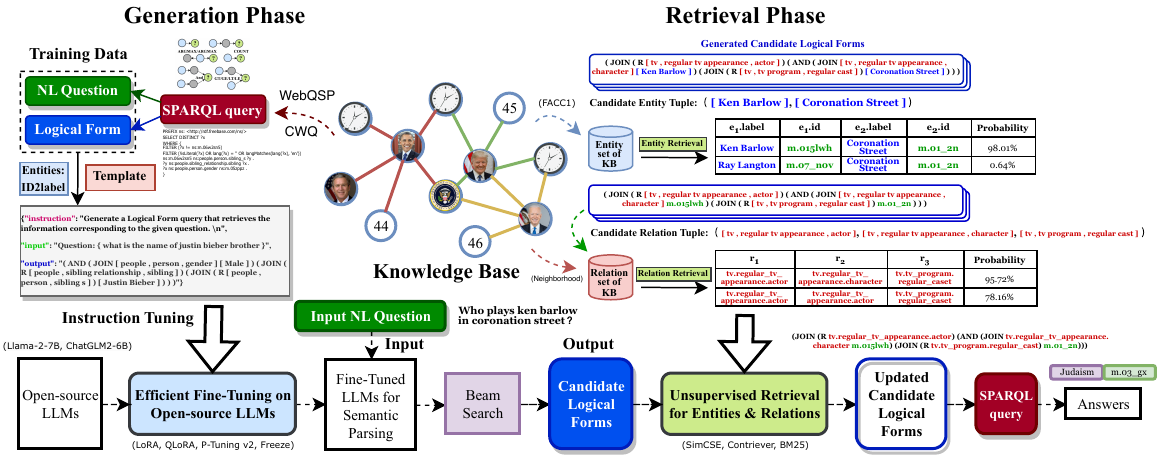}
\caption{The overview of ChatKBQA framework for generate-then-retrieve KBQA method with fine-tuned LLMs and unsupervised retrieval for entities and relations in candidate logical forms.}
\label{f2}
\end{figure*}

\section{Methodology}
In this section, we first present an overview of the ChatKBQA framework as shown in Figure~\ref{f2}, and introduce efficient fine-tuning on large language models (LLMs), logical form generation by fine-tuned LLMs, unsupervised retrieval for entities and relations, and interpretable query execution.
\subsection{Overview of ChatKBQA}
ChatKBQA is a generate-then-retrieve KBQA framework with fine-tuned LLMs. First, the ChatKBQA framework needs to efficiently fine-tune an open-source LLM based on the (natural language question, logical form) pairs in the KBQA dataset by instruction tuning. The fine-tuned LLM is then used to convert the new natural language questions to according candidate logical forms by semantic parsing. Then, ChatKBQA retrieves the entities and relations in these logical forms at the phrase level, and searches for the logical forms that can be executed against KB after being converted to SPARQL. Finally, the converted SPARQL generates the final set of answers, resulting in interpretable and knowledge-required responses to natural language questions.

\subsection{Efficient Fine-Tuning on LLMs }
To construct the instruction fine-tuning training data, ChatKBQA first converts the SPARQL corresponding to the natural language questions of the train set in the KBQA dataset into equivalent logical forms and then replaces the entity IDs (e.g., "\texttt{m.06w2sn5}") in these logical forms with the corresponding entity tags (e.g., "\texttt{[ Justin Bieber ]}"), to let LLMs understand entity labels better than meaningless entity IDs. We then combine the natural language question (e.g. "\texttt{What is the name of Justin Bieber's brother?}") and the processed corresponding logical form (e.g. "\texttt{(AND (JOIN [ people, person, gender ] [ Male ]) (JOIN (R [ people, sibling relationship, sibling ]) (JOIN (R [ people, person, sibling s ]) [ Justin Bieber ])))}") as "input" and "output" respectively, and add "instruction" as "\texttt{Generate a Logical Form query that retrieves the information corresponding to the given question.}" constitutes the instruction fine-tuning training data for LLMs.

ChatKBQA employs Parameter Efficient Fine-Tuning (PEFT)~\citep{PEFT} techniques including various efficient fine-tuning methods, such as LoRA~\citep{LoRA}, QLoRA~\citep{QLoRA}, P-tuning v2~\citep{P-Tuningv2}, and Freeze~\citep{Freeze}, to minimize the cost of fine-tuning LLMs with a large number of parameters. ChatKBQA can switch between all the above fine-tuning methods as well as open-source LLMs, such as Llama-2-7B~\citep{Llama2} and ChatGLM2-6B~\citep{GLM}.

\subsection{Logical Form Generation by LLMs}
Through fine-tuning, the LLMs have acquired expertise in semantic parsing, enabling them to convert natural language questions into logical forms. We apply the fine-tuned LLMs to perform semantic parsing on the new questions in the test set and observe that approximately 63\% of the samples match the ground truth logical forms exactly. When employing beam search, the set of candidate logical forms $\mathcal{C}$ generated by our LLMs includes approximately 74\% of the instances with correct logical forms, indicating that fine-tuned LLMs possess effective learning and parsing abilities for semantic parsing tasks. In addition, by replacing the entities and relations in the candidate logical forms with "\texttt{[]}" (for example, "\texttt{(AND (JOIN [] []) (JOIN (R []) (JOIN (R []) [])))}"), more than 91\% of the samples contain the candidate skeleton. Hence, the next step involves retrieving the entities and relations in the logical form with the corresponding ones from the KB to enhance performance further.

\subsection{Unsupervised Retrieval for Ents and Rels}
Due to the strong generative capabilities of fine-tuned LLMs for logical form skeletons, we employ an unsupervised retrieval approach during the retrieval phase. This method involves subjecting the entities and relations in the candidate logical forms to phrase-level semantic retrieval and replacement. The result is a final logical form that can be executed as a SPARQL query against the KB.

\vspace{-2mm}
\setlength{\algomargin}{6pt}
\SetInd{0em}{1.2em}
\begin{algorithm}[h] 
\small
% \raggedright
% \SetKwProg{Fn}{Function}{}{}
\SetKwFunction{TopKwithThreshold}{TopKwithThreshold}
\SetKwFunction{PermuteByEntity}{PermuteByEntity}
\SetKwFunction{PermuteByRelation}{PermuteByRelation}
\SetKwFunction{SimiEntities}{SimiEntities}
\SetKwFunction{SimiRelations}{SimiRelations}
\SetKwFunction{Neighborhood}{Neighborhood}
% \SetKwFunction{Convert}{Convert}
% \SetKwFunction{Execute}{Execute}
% \SetKwFunction{Sp}{Sp}
\SetKw{continue}{continue}
\SetKw{return}{return}
\SetKwInOut{Input}{Input}
\SetKwInOut{Output}{Output}
\Input{Candidate logical forms generated from LLM $\mathcal{C}$, top-k threshold $k_e, k_r, k_1, k_2$, probability threshold $t_e, t_r, t_1, t_2$, the entity set of KB $\mathcal{E}$}
\Output{The equivalent SPARQL query $q$}
$\mathcal{C}^\prime\ \leftarrow\ \emptyset$\ \ForEach{$F \in \mathcal{C}$}{
    \ForEach{$e \in F$}{
        $e_{list} \leftarrow \emptyset$\ \ForEach{$e^\prime \in \mathcal{E}$}{
            $s_e \leftarrow \SimiEntities(e, e^\prime)$\;
            $e_{list}.append((e^\prime, s_e))$\;
        }
        $e_{list} \leftarrow \TopKwithThreshold(e_{list}, k_e, t_e)$\;
        $F.attach(e_{list})$\;
    }
    $F_{list} \leftarrow \PermuteByEntity(F)$\;
    $\mathcal{C}^\prime.append(\TopKwithThreshold(F_{list}, k_1, t_1))$\;
}
$\mathcal{C}^{\prime\prime}\ \leftarrow\ \emptyset$\ \ForEach{$F \in \mathcal{C}^\prime$}{
    \ForEach{$e \in F$}{
        $r_{list} \leftarrow \emptyset$\ \ForEach{$r \in \Neighborhood(\mathcal{E}_F)$}{
            $s_r \leftarrow \SimiRelations(r, r^\prime)$\;
            $r_{list}.append((r^\prime, s_r))$\;
        }
        $r_{list} \leftarrow \TopKwithThreshold(r_{list}, k_r, t_r)$\;
        $F.attach(r_{list})$\;
    }
    $F_{list} \leftarrow \PermuteByRelation(F)$\;
    $\mathcal{C}^{\prime\prime}.append(\TopKwithThreshold(F_{list}, k_2, t_2))$\;
}
\ForEach{$F \in \mathcal{C}^{\prime\prime}$}{
    $q=\Convert(F)$\ \If{$q$ is valid to execute}{
        \return $q$\;
    }
}
$\return\ \emptyset$\;
\caption{Unsupervised Retrieval}
\label{algo:er_retrival}
\end{algorithm}
\vspace{-2mm}
Specifically, as shown in the Algorithm~\ref{algo:er_retrival}, the input is the generated candidate logical form list $\mathcal{C}$, and we traverse each of these logical forms $F$ in order. First, we perform the entity retrieval. For each entity $e$ in $F$, we compute the similarity $s_e \leftarrow \SimiEntities(e, e^\prime)$ with the label of each entity $e^\prime$ in the knowledge base $\mathcal{K}$ entity set $\mathcal{E}$. We sort the retrieved entities based on the similarities, take the top $k_e$ and greater than the threshold $t_e$ to get the retrieval result for that entity $e_{list} \leftarrow \TopKwithThreshold(e_{list}, k_e, t_e)$. Function $\PermuteByEntity$ performs permutation on the retrieved entities at each position, and we get the result $F_{list}$ after entity retrieval. Based on probabilities in $F_{list}$, we take top $k_1$ and greater than threshold $t_1$ to get a new candidate logical form list $\mathcal{C}^\prime.append(\TopKwithThreshold(F_{list}, k _1, t_1))$.

Then, we perform the relation retrieval. Similar to entity retrieval, for each relation $r$ in $F \in \mathcal{C}^\prime$, we compute the similarity $s_r \leftarrow \SimiRelations(r, r^\prime)$ with each candidate relation $r^\prime$ according to the neighborhood of entity set of the logical form $\mathcal{E}_{F}$. We also sort the retrieved relations according to the similarities, take the top $k_r$ and greater than the threshold $t_r$ to get the retrieval result $r_{list} \leftarrow \TopKwithThreshold(r_{list}, k_r, t_r)$. By permuting the retrieval results of the relations at each position, we get the result $F_{list}$ after relation retrieval and then take top $k_2$ and greater than the threshold $t_2$ to get a new list of candidate logical forms $\mathcal{C}^{\prime\prime}.append(\TopKwithThreshold(F_{list}, k_2, t_2))$.

Given a query, unsupervised retrieval methods such as SimCSE~\citep{SimCSE}, Contriever~\citep{Contriever}, and BM25~\citep{BM25}, require no additional training to identify the top k most semantically similar candidates from the set of retrieved answers. ChatKBQA can switch between all the above unsupervised retrieval methods. We also discuss the retrieval complexity in Appendix~\ref{complex}.

\subsection{Interpretable Query Execution}
After retrieval, we get a final candidate logical form list $\mathcal{C}^{\prime\prime}$, which we sequentially iterate through the logical form $F \in \mathcal{C}^{\prime\prime}$ and convert to the equivalent of the SPARQL query $q=\Convert(F)$. When the first $q$ that can be executed against KB $\mathcal{K}$ is found, we execute to get the final answer set $A=\Execute(q|\mathcal{K})$. With this approach, we can also get a complete reasoning path for natural language questions based on SPARQL query with good interpretability. To summarize, ChatKBQA proposes a thought taking both the advantages of using LLMs to do natural language semantic parsing for graph query generation and calling external KBs to interpretably reason with queries.

% , which we name Graph Query of Thoughts (GQoT), a promising LLM+KG combination paradigm to better utilize the external knowledge, improve Q\&A's interpretability, and avoid LLM's hallucinations.

\section{Experiments}
This section presents the experimental setup, results, and analysis. We answer the following research questions (RQs):
\textbf{RQ1}: Does ChatKBQA outperform other KBQA methods?
\textbf{RQ2}: Does the main components of ChatKBQA work?
\textbf{RQ3}: Why use Generate-then-Retrieve method instead of Retrieve-then-Generate method?
\textbf{RQ4}: Why use fine-tuned open-source LLMs instead of calling ChatGPT or training traditional T5 models?
\textbf{RQ5}: Does Generate-then-Retrieve method improve retrieval efficiency?
\textbf{RQ6}: Does ChatKBQA has plug-and-play characteristics?
% \textbf{RQ7}: How about error analysis?

% \textbf{RQ7}: How does ChatKBQA perform in specific case study?

\subsection{Experimental Setup}
\textbf{\ \ \ \ Datasets. } All experiments are conducted on two standard KBQA datasets: WebQuestionsSP (WebQSP)~\citep{STAGG} containing 4,737 natural language questions with SPARQL queries and ComplexWebQuestions (CWQ)~\citep{CWQ} containing 34,689 natural language questions with SPARQL queries. Both datasets are based on Freebase~\citep{Freebase} KB. More details of datasets are in Appendix~\ref{dataset}.

\textbf{Baselines. } We compare ChatKBQA with numerous KBQA baseline methods, including IR-based methods~\citep{KV-Mem}, SP-based methods~\citep{STAGG, CBR-KBQA}, and LLM-based methods~\citep{StructGPT} in Section~\ref{s2}. Details of more baselines are in Appendix~\ref{baseline}.
% \textbf{Ablations. } To evaluate the significance of ChatKBQA's three main components, logical form generation by fine-tuned LLMs, entity retrieval (ER) and relation retrieval (RR), we obtain three simplified variants by removing ER or RR component from the framework (\textbf{ChatKBQA w/o ER}, \textbf{ChatKBQA w/o RR}, and \textbf{ChatKBQA w/o ER,RR}) for comparison.

\textbf{Evaluation Metrics. }
Following previous work~\citep{TIARA,DECAF}, we use $F_1$ score, Hits@1, and Accuracy (Acc) to denote coverage of all the answers, single top-ranked answer, and strict exact-match accuracy, respectively.

\textbf{Hyperparameters and Enviroment. }
We fine-tune LLMs 100 epochs on WebQSP and 10 epochs on CWQ with batch size 4 and learning rate 5e-5, detailed in Apendix~\ref{hyper}. All experiments were done on a single NVIDIA A40 GPU (48GB), with results averaged from five randomly seeded experiments. 
% Appendix~\ref{hyper} shows the optimal hyperparameter settings for ChatKBQA.  Appendix~\ref{train} shows training details.
\begin{table}[h!t]
% \scriptsize
% \footnotesize
\small
\setlength{\tabcolsep}{0.6mm}{
\begin{tabular}{lcccccc}
\toprule
                                 & \multicolumn{3}{c}{\textbf{WebQSP}}                                                               & \multicolumn{3}{c}{\textbf{CWQ}}                                             \\ \cline{2-7} 
\multirow{-2}{*}{\textbf{Model}} & \textbf{F1}   & \textbf{Hits@1}  & \multicolumn{1}{c|}{\textbf{Acc}}  & \textbf{F1}   & \textbf{Hits@1} & \textbf{Acc}  \\ \hline
\multicolumn{7}{c}{\textit{IR-based KBQA Methods}}                                                                \\ \hline
KV-Mem                           & 34.5          & 46.7                                         & \multicolumn{1}{c|}{-}             & 15.7          & 21.1                                         & -             \\
PullNet                          & -             & 68.1                                         & \multicolumn{1}{c|}{-}             & -             & 47.2                                         & -             \\
EmbedKGQA*                       & -             & 66.6                                         & \multicolumn{1}{c|}{-}             & -             & 44.7                                         & -             \\
NSM+h*                          & 67.4          & 74.3                                         & \multicolumn{1}{c|}{-}             & 44.0          & 48.8                                         & -             \\
TransferNet                     & -             & 71.4                                         & \multicolumn{1}{c|}{-}             & -             & 48.6                                         & -             \\
Subgraph Retrieval*              & 64.1          & 69.5                                         & \multicolumn{1}{c|}{-}             & 47.1          & 50.2                                         & -             \\ \hline
\multicolumn{7}{c}{\textit{SP-based KBQA Methods (step-wise)}}                                                                \\ \hline
STAGG                           & 71.7          & -                                            & \multicolumn{1}{c|}{63.9}          & -             & -                                            & -             \\
UHop                             & 68.5          & -                                            & \multicolumn{1}{c|}{-}             & 29.8          & -                                            & -             \\
Topic Units                     & 67.9          & 68.2                                         & \multicolumn{1}{c|}{-}             & 36.5          & 39.3                                         & -             \\
QGG                             & 74.0          & 73.0                                         & \multicolumn{1}{c|}{-}             & 40.4          & 44.1                                         & -             \\
UniKGQA*                         & 72.2          & 77.2                                         & \multicolumn{1}{c|}{-}             & 49.4          & 51.2                                         & -             \\ \hline
\multicolumn{7}{c}{\textit{SP-based KBQA Methods (seq2seq)}}                                                                \\ \hline
CBR-KBQA                         & 72.8          & -                                            & \multicolumn{1}{c|}{69.9}          & 70.0          & 70.4                                         & 67.1          \\ 
RnG-KBQA                         & 75.6          & -                                            & \multicolumn{1}{c|}{71.1}          & -             & -                                            & -             \\
Program Transfer*                & 76.5          & 74.6                                         & \multicolumn{1}{c|}{-}             & 58.7          & 58.1                                         & -             \\
TIARA*                           & 78.9          & 75.2                                         & \multicolumn{1}{c|}{-}             & -             & -                                            & -             \\
GMT-KBQA                         & 76.6          & -                                            & \multicolumn{1}{c|}{73.1}          & 77.0          & -                                            & 72.2          \\
UnifiedSKG                       & 73.9          & -                                            & \multicolumn{1}{c|}{-}             & 68.8          & -                                            & -             \\
DecAF                            & 78.8          & 82.1                                         & \multicolumn{1}{c|}{-}             & -             & 70.4                                         & -             \\
FC-KBQA                          & 76.9          & -                                            & \multicolumn{1}{c|}{-}             & 56.4          & -                                            & -             \\ \hline
\multicolumn{7}{c}{\textit{LLM-based KBQA Methods}}                                                                \\ \hline
StructGPT*                       & 72.6          & -                                            & \multicolumn{1}{c|}{-}             & -             & -                                            & -             \\
Pangu                            & 79.6          & -                                            & \multicolumn{1}{c|}{-}             & -             & -                                            & -             \\ 
ToG*                         & -          & 82.6                                            & \multicolumn{1}{c|}{-}             & -         & 69.5                                            & -             \\ \hline\hline
ChatKBQA (ours)                        & 79.8          & 83.2                                         & \multicolumn{1}{c|}{73.8}          & 77.8          & 82.7                                         & 73.3          \\
ChatKBQA* (ours)                      & \textbf{83.5} & \textbf{86.4}                                & \multicolumn{1}{c|}{\textbf{77.8}} & \textbf{81.3} & \textbf{86.0}                                & \textbf{76.8} \\ \bottomrule
\end{tabular}}
\caption{\label{main}
KBQA comparison of ChatKBQA with other baselines on WebQSP and CWQ datasets. * denotes using oracle entity linking annotations. The results of the models are mainly taken from their original paper. For our proposed ChatKBQA framework, we display the results of the best setup on WebQSP and CWQ, respectively. The best results in each metric are in \textbf{bold}. 
}
\end{table}

\subsection{Main Results (RQ1)}
For the KBQA task, Table~\ref{main} lists the experimental results for our proposed generate-then-retrieve ChatKBQA framework, with the best setup of LoRA~\citep{LoRA} fine-tuning Llama-2-7B~\citep{Llama2} (beam size = 15) on WebQSP, Llama-2-13B~\citep{Llama2} (beam size = 8) on CWQ, with SimCSE~\citep{SimCSE} for unsupervised retrieval, and other baseline models. We can see that ChatKBQA has shown improvement over all existing KBQA methods on both WebQSP and CWQ datasets. The $F_1$ score, Hits@1, and Acc have improved by about 4, 4, and 4 percentage points on WebQSP and about 4, 16, and 4 percentage points on CWQ, respectively, compared to the previous best results, which reflects ChatKBQA's superior KBQA capability to reach the new state-of-the-art performance.
% \begin{figure}[h!t]
%     \vspace{-3mm}
%     \centering
%     \subfigure[\label{a}]{
%         \includegraphics[width=0.47\textwidth, height=0.36\textwidth]{output1.pdf}
%     }
%     \subfigure[\label{b}]{
%         \includegraphics[width=0.40\textwidth, height=0.36\textwidth]{output2.pdf}
%     }
% 	\caption{(a) Ablation study in ChatKBQA generation phase. (b) Ablation study in ChatKBQA retrieval phase.}
% 	\label{f3}
% \end{figure}

\begin{figure*}[h!t]
    % \vspace{-3mm}
    \centering
    \subfigure[\label{a}]{
        \includegraphics[width=0.235\textwidth, height=0.18\textwidth]{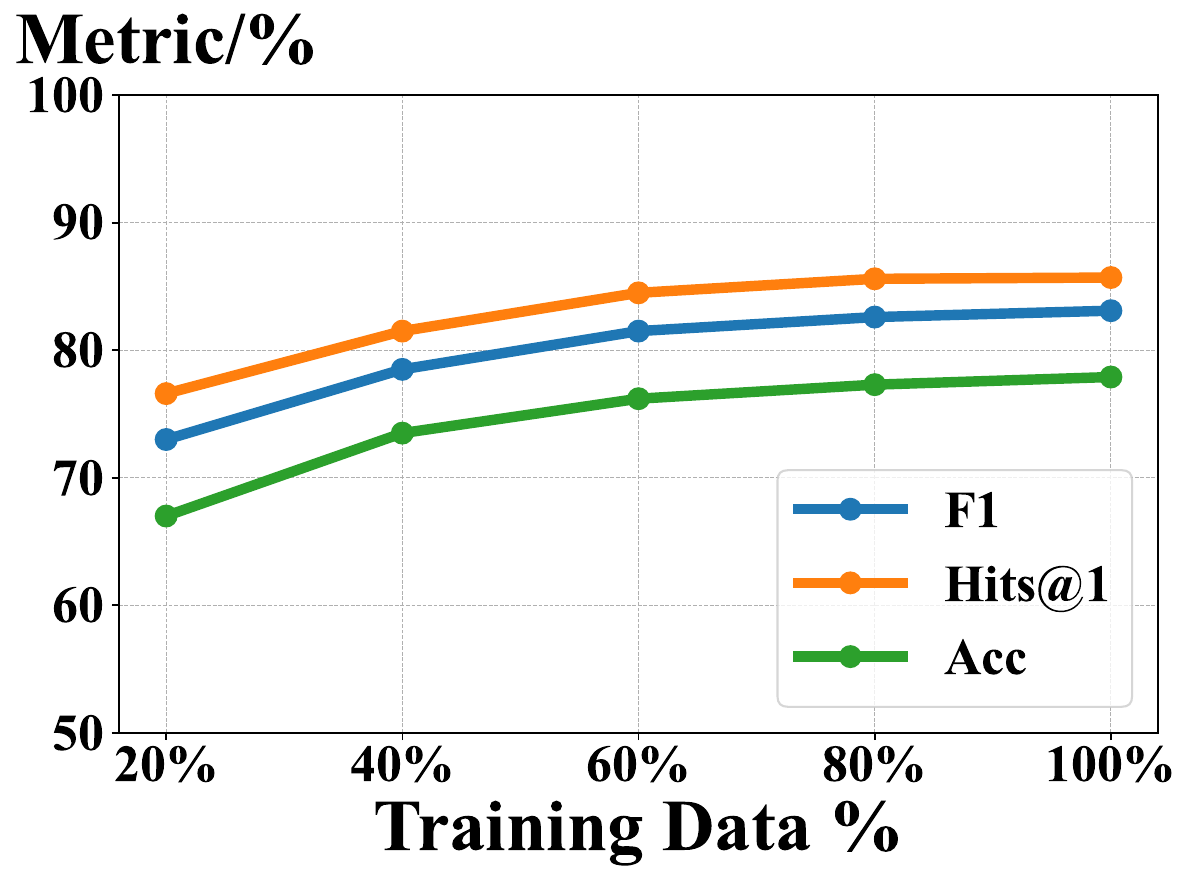}
    }
    \subfigure[\label{b}]{
        \includegraphics[width=0.20\textwidth, height=0.18\textwidth]{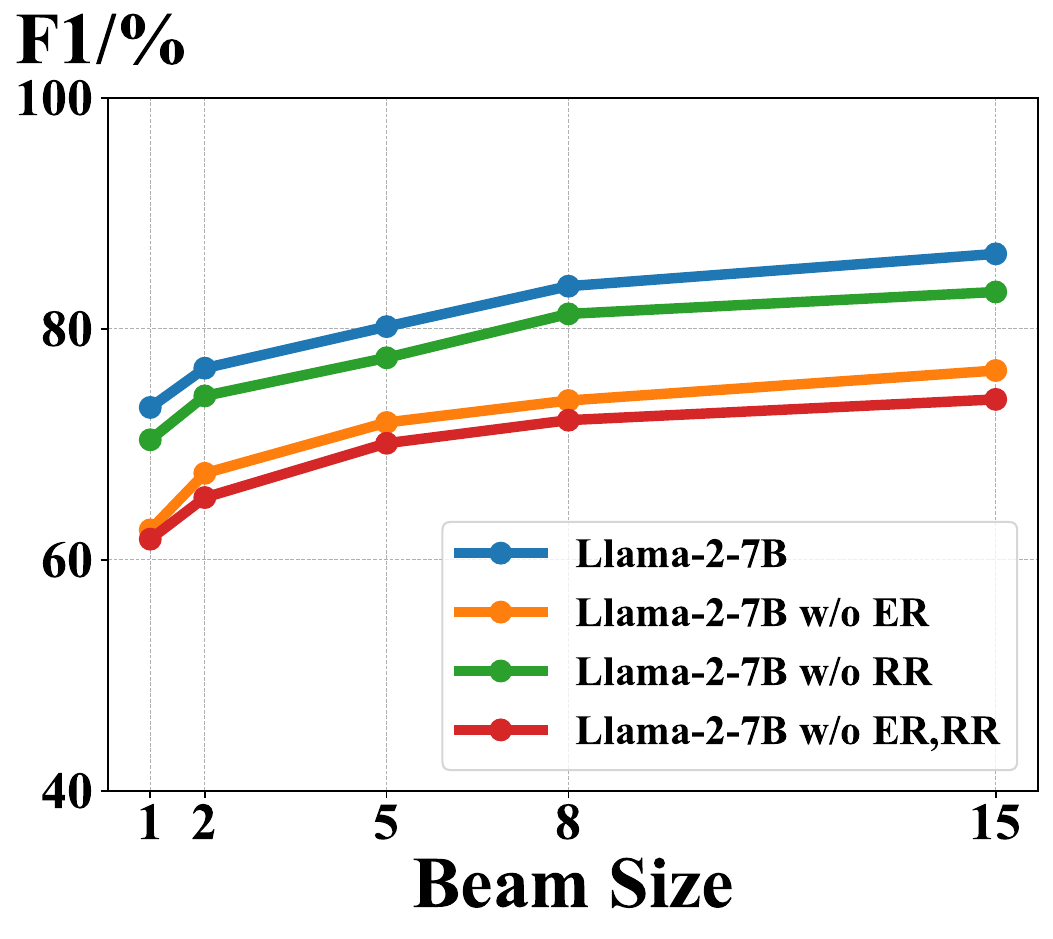}
    }
    %\vfill
    \subfigure[\label{c}]{
        \includegraphics[width=0.245\textwidth, height=0.18\textwidth]{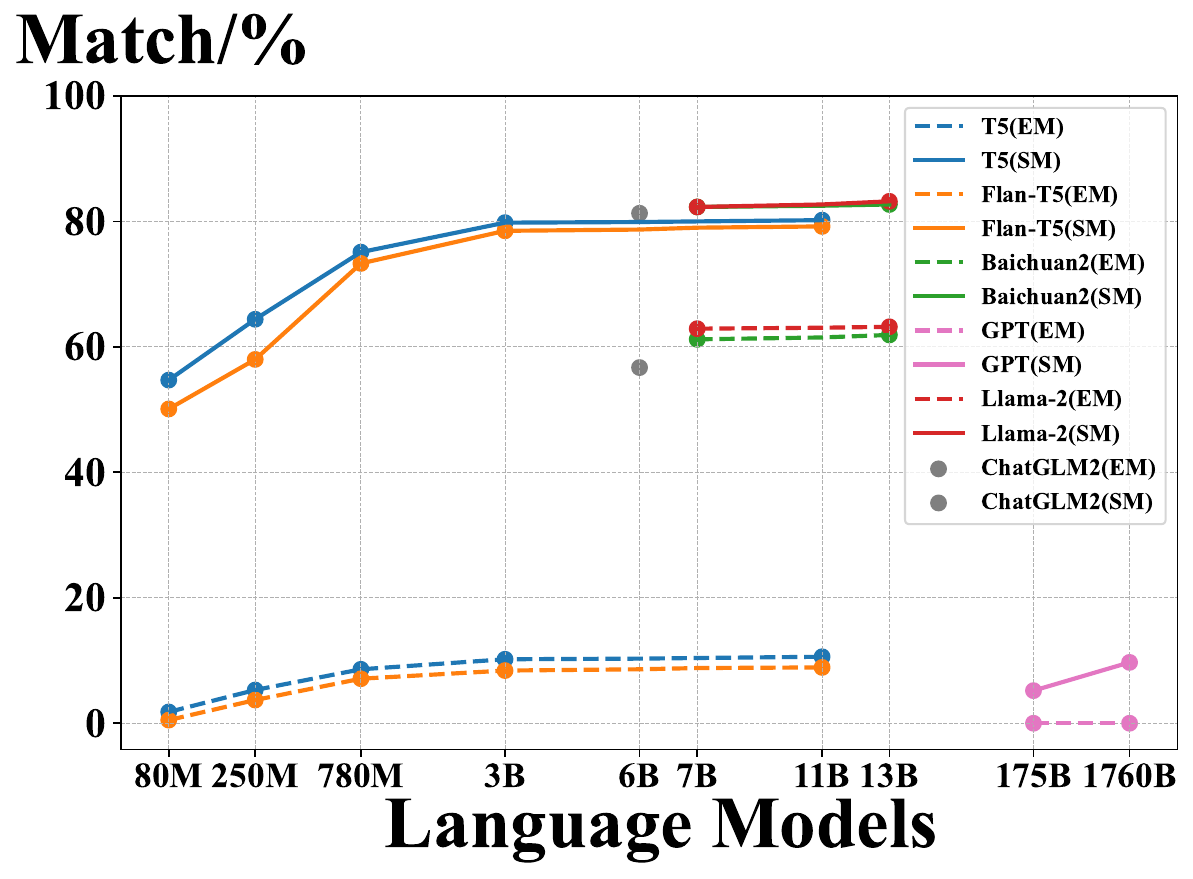}
    }
    \subfigure[\label{d}]{
        \includegraphics[width=0.25\textwidth, height=0.18\textwidth]{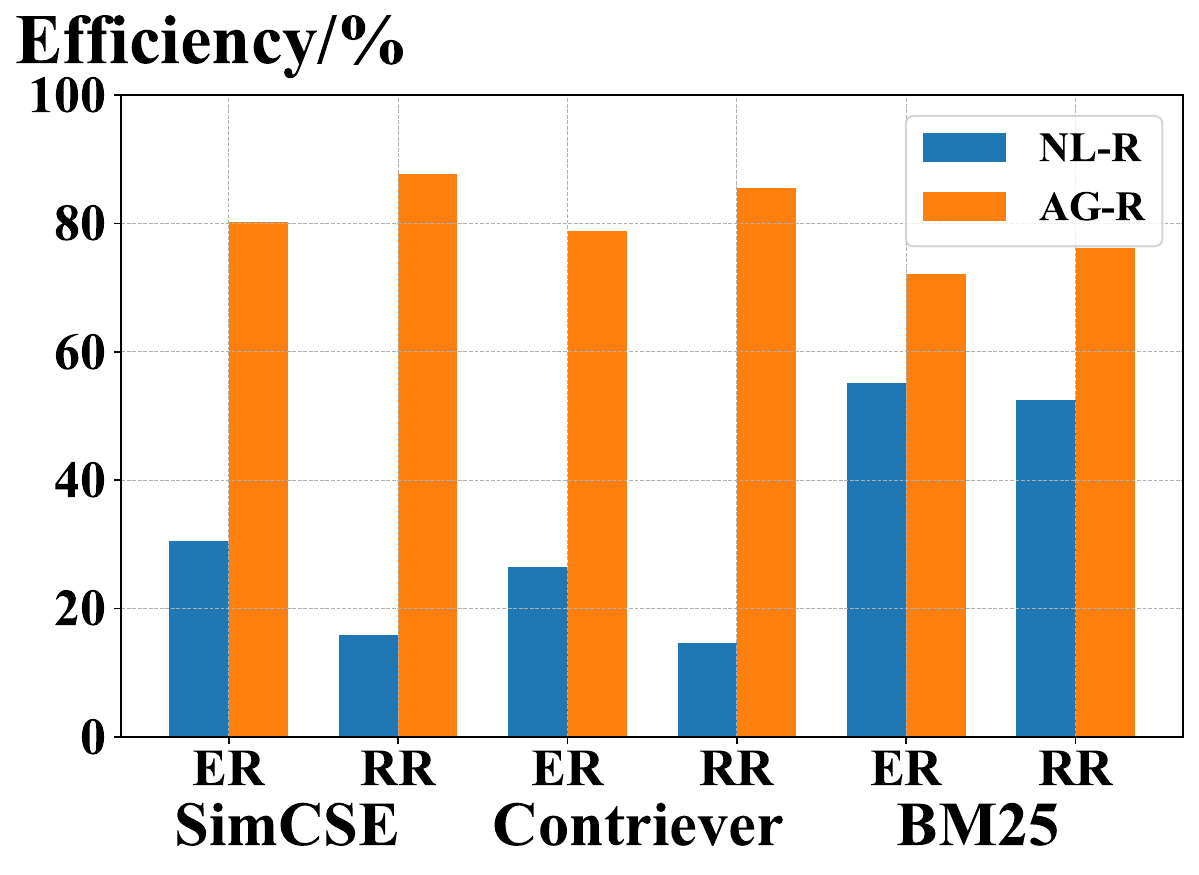}
    }
	\caption{(a) Ablation study in ChatKBQA generation phase. (b) Ablation study in ChatKBQA retrieval phase. (c) Comparison with other language models in the generation phase. (d) Comparison of retrieval efficiency between retrieval from nature language questions (NL-R) and generated logical forms (AG-R) in the retrieval phase.}
	\label{f3}
\end{figure*}

\subsection{Ablation Study (RQ2)}

To validate the effectiveness of the generation and retrieval phases of ChatKBQA, we ablate the two phases separately. 
For the generation phase, we use 20\%, 40\%, 60\%, and 80\% of the training data for fine-tuning versus full training set fine-tuning. 
For the retrieval phase, to validate entity retrieval (ER) and relation retrieval (RR) separately, we remove ER or RR from the framework and obtain three simplified variants for comparison.
% (\textbf{ChatKBQA w/o ER}, \textbf{ChatKBQA w/o RR} and \textbf{ChatKBQA w/o ER,RR}) at four different beam sizes for comparison. 
% \begin{figure}[H]
%     \centering
%     \begin{minipage}{0.24\textwidth}
%         \centering
%         \includegraphics[width=\textwidth, height=0.80\textwidth]{output1.pdf}
%     \end{minipage}\hfill
%     \begin{minipage}{0.23\textwidth}
%         \centering
%         \includegraphics[width=\textwidth, height=0.83\textwidth]{output2.pdf}
%     \end{minipage}
%     \caption{Ablation study in ChatKBQA generation phase (left) to verify the effectiveness of LLM's Fine-tuning. Ablation study in ChatKBQA retrieval phase (right) to verify the effectiveness of Beam Search, Entity Retrieval (ER), and Relation Retrieval (RR).}
%      \label{a}
% \end{figure}

\textbf{Effectiveness of LLM's Fine-tuning. }
As shown in Figure~\ref{a}, the performance of KBQA gets better as the training volume increases, proving the effectiveness of fine-tuning. We also observe that the F1 score has exceeded 70\% when only using 20\% training data to fine-tune, which indicates that the fine-tuned LLMs are also effective at learning from a limited dataset. As shown in Figure~\ref{b}, we also utilize beam search to improve the generation performance, detailed in Appendix~\ref{beam}.
% However, as the training data increases, the metrics also show improvement but plateaus, possibly due to the complexity or ambiguity of some questions in the dataset that cannot be resolved with additional training data alone.

\textbf{Effectiveness of Entity Retrieval (ER). }
As shown in Figure~\ref{b}, ER improves about 15 percentage points on average over no oracle entity linking in the F1 score at different beam sizes. This is because, after LLM's fine-tuning, the generated logical forms contain entities unseen in the train set, which can be further aligned to KB after retrieving the entities from the KB entity set.

\textbf{Effectiveness of Relation Retrieval (RR). }
As shown in Figure~\ref{b}, RR enhances the F1 score by an average of 5\% across various beam sizes in ablation experiments. Although relations are rarely directly present in natural language questions, the number of thousand-level relations in the KB is still small compared to the tens of millions of entities, and the LLM perceives relational information well during fine-tuning. Thus, RR does not improve performance as much as ER, but combined with ER, RR makes KBQA perform at its best.

\subsection{Generate-then-Retrieve Or Retrieve-then-Generate (RQ3)}
To verify that our proposed LLM-based Generate-then-Retrieve method is better than previous Retrieve-then-Generate methods, we add Top1, Top2, Top5, and Top10 retrieval knowledge fragments obtained in DecAF~\citep{DECAF}  to the instruction, respectively, compared with the fine-tuning of Llama-2-7B without retrieval. 
\begin{table}[h!t]
\centering
\small
\centering
\setlength{\tabcolsep}{0.05mm}{
\begin{tabular}{lcccc}
\toprule
\multirow{2}{*}{\textbf{Fine-tuning Settings}} & \multicolumn{4}{c}{\textbf{WebQSP}}                               \\ \cline{2-5} 
                                      & \textbf{Max Token↓} & \textbf{EM↑} \% & \textbf{BM↑} \% & \textbf{SM↑} \% \\ \hline
Llama-2-7B w/o R             & \textbf{512}     & \textbf{63.5}     & \textbf{74.7 }         & \textbf{91.1}              \\ \hline
Llama-2-7B w Top1 R          & 612     & 58.5     & 72.3          & 88.4              \\
Llama-2-7B w Top2 R          & 712     & 59.7     & 73.6          & 89.0              \\
Llama-2-7B w Top5 R          & 1012    & 55.6     & 68.3          & 85.3              \\
Llama-2-7B w Top10 R         & 2012    & 53.1     & 67.9          & 84.8              \\ \bottomrule
\end{tabular}
}
\caption{\label{t2}
Comparison of whether or not utilizing retrieval results before fine-tuning Llama-2-7B for logical form generation in ChatKBQA. 
}
\vspace{-2mm}
\end{table}

As shown in Table~\ref{t2}, we find that without retrieval is better than with retrieval in the logical form generation in terms of extract match ratio (EM), match after beam search ratio (BM), and skeleton match ratio (SM), because the information obtained from retrieval will \textbf{have erroneous interfering information} and \textbf{increase Max Token of instruction}, which leads to catastrophic forgetting of the original problem for LLMs and increases the difficulty of training. At the same time, we observe that Llama-2-7B fine-tuning without retrieval achieves a BM of 74.7\% and SM hits 91.1\%, with good performance because of LLM's well-learned schema of entities and relations, which provides the basis for the retrieval after generation.

\subsection{Comparison with ChatGPT and T5 in Generation Phase (RQ4)}

To illustrate why ChatKBQA chooses to fine-tune open-source generative LLMs such as Llama-2-7B and ChatGLM2-6B, we replace the LLMs in the generation phase with ChatGPT and GPT-4~\citep{GPT4} with API call in a zero-shot setting, T5~\citep{T5} and Flan-T5~\citep{Flan-T5} with seq2seq training, respectively, and observe their results in Extract Match (EM) and Skeleton Match (SM) results without beam search. 
% \begin{figure}[H]
% % \centering
% \includegraphics[width=0.45\textwidth, height=0.30\textwidth]{output4.pdf}
% \caption{Comparison of extract match (EM) and skeleton match (SM) results of logical form generation with other language models in generation phase.}
% \label{c}
% \end{figure}

\textbf{Comparison with zero-shot ChatGPT. }
As shown in Figure~\ref{c}, ChatGPT and GPT-4, although having large parametric quantities, cannot generate standard logical forms well because they aren't open-source to be fine-tuned. They can generate the SPARQL language, but it is challenging to build the correct query skeleton, entities, and relations because they cannot perceive the complex structure of the external KB well through designing prompts in limited context length.

\textbf{Comparison with fine-tuned T5 \& Flan-T5. }
While T5 and Flan-T5 can capture the skeletons well after fine-tuning, the EM is only about 10\%, which is much worse than the 63\% of Llama-2-7B, and therefore does not guarantee subsequent unsupervised entity and relation retrieval. Fine-tuned open-source LLMs such as Llama-2-7B~\citep{Llama2} and ChatGLM2-6B~\citep{GLM} show stronger semantic parsing ability than models such as T5 and ChatGPT and can generate higher-quality logical forms in both EM \& SM.

\subsection{Analysis of Efficiency of Retrieval in Retrieval Phase (RQ5)}

To embody the Generate-then-Retrieve method improving the efficiency of retrieval, we compare entity retrieval (ER) and retrieval (RR) after logical form generation (AG-R) with traditional retrieval from natural language questions (NL-R). We define the efficiency of retrieval as the average similarity ranging [0,1] between the text to be retrieved and the set of retrieved answers, which is scored by different retrieval models. Note that the BM25 score needs to be mapped to the similarity range of [0,1]. 
% \begin{figure}[H]
% % \centering
% \includegraphics[width=0.45\textwidth, height=0.30\textwidth]{output3.pdf}
% \caption{Comparison of retreival efficiency between retrieval from nature language questions (NL-R) and generated logical forms (AG-R) in retrieval phase.}
% \label{d}
% \end{figure}

\textbf{Efficiency gains in both ER \& RR. }
As Figure~\ref{d} shows, all three retrieval methods SimCSE~\citep{SimCSE}, Contriever~\citep{Contriever}, and BM25~\citep{BM25} consider AG-R to be more efficient than NL-R for both ER and RR. This is due to NL-R still needs to determine the boundaries of the entities or relations. However, this step has been completed in AG-R after LLM generates the logical forms. 

\textbf{RR has more efficiency gains than ER. }
Moreover, although the generated logical form has fewer kinds of relations than entities in general, the relations generally exist implicitly in natural language questions. Thus, relations are more difficult to determine the boundaries than entities in natural language questions, and the generation of logical forms with the help of fine-tuned LLMs can help us to better determine the boundaries of relations, resulting in a more significant improvement in the efficiency of RR over ER.

\subsection{Plug-and-Play Characteristics (RQ6)}

\begin{table}[t]
% \scriptsize
\small
\centering
\setlength{\tabcolsep}{0.7mm}{
\begin{tabular}{cccccc}
\toprule
\multicolumn{3}{c}{\textbf{ChatKBQA Framework}}                    & \multicolumn{3}{c}{\textbf{WebQSP}}                                                                        \\ \hline
\textbf{LLMs} & \textbf{Tuning} & \textbf{Retrieval} & \textbf{F1}                  & \textbf{Hits@1} & \textbf{Acc}                 \\ \hline
Baichuan2-7B  & LoRA                   & SimCSE                    & 79.1                         & 81.5                                         & 74.1                         \\
Baichuan2-13B & LoRA                   & SimCSE                    & 79.4                         & 82.1                                         & 74.4                         \\
ChatGLM2-6B   & LoRA                   & SimCSE                    & 79.8                         & 82.7                                         & 74.5                         \\
Llama-2-7B    & LoRA                   & SimCSE                    & 80.0                         & 82.4                                         & 75.2                         \\
\textbf{Llama-2-13B}   & \textbf{LoRA}                   & \textbf{SimCSE}                    & \textbf{82.6}                & \textbf{85.2}                                & \textbf{77.5}                \\ \hline
Llama-2-13B   & QLoRA                  & SimCSE                    & 81.9 & 85.0                 & 76.9 \\
ChatGLM2-6B   & P-Tuning v2            & SimCSE                    & 74.6 & 77.8                 & 70.6 \\
Llama-2-13B   & Freeze                 & SimCSE                    & 81.7 & 84.7                 & 76.8 \\ \hline
Llama-2-13B   & LoRA                   & Contriever                & 81.5 & 83.6                 & 76.8 \\
Llama-2-13B   & LoRA                   & BM25                      & 79.8 & 80.5                 & 72.7 \\ \bottomrule
\end{tabular}
}
\caption{\label{framework}
Plug-and-play performance comparison of ChatKBQA framework for replacing LLMs, tuning methods, and unsupervised retrieval methods, respectively, with the beam size all set as 8.
}
\end{table}

ChatKBQA is a KBQA framework based on LLMs with plug-and-play characteristics that can flexibly replace three parts: LLM, efficient tuning method, and unsupervised retrieval method. We choose Llama-2-13B~\citep{Llama2} for LLM, LoRA~\citep{LoRA} for the tuning method, and SimCSE~\citep{SimCSE} for the retrieval method as the basic variant, setting the beam size for all variants to 8 for comparison. 

We replace Baichuan2-7B~\citep{Baichuan2}, Baichuan2-13B~\citep{Baichuan2}, ChatGLM2-6B~\citep{GLM}, Llama-2-7B~\citep{Llama2} \textbf{in the LLM part}, QLoRA~\citep{QLoRA}, P-Tuning v2~\citep{P-Tuningv2}, Freeze~\citep{Freeze} \textbf{in the tuning part}, and Contriever~\citep{Contriever}, BM25~\citep{BM25} \textbf{in the retrieval part}. Benefiting from the plug-and-play characteristics of the ChatKBQA framework, as the LLMs and the methods of tuning and retrieval are upgraded, the KBQA task will be solved better with good flexibility and extensibility. More details are in Appendix~\ref{play}.

% \subsection{Case Study (RQ7)}

% For the KBQA task, Table~\ref{t4} lists the experimental results for our proposed generate-then-retrieve ChatKBQA framework, with the best setup of using Llama-2-13B~\cite{Llama2} for LoRA~\cite{LoRA} fine-tuning and SimCSE~\cite{SimCSE} for unsupervised retrieval, and other baseline models. We can see that ChatKBQA is a significant improvement over all existing KBQA methods on both WebQSP and CWQ datasets. The $F_1$ scores, Hits@1, and Acc are improved by about 3, 3, and 2 percentage points on WebQSP and about 3, 15, and 4 percentage points on CWQ, respectively, compared to the previous best results, which reflects ChatKBQA's superior KBQA capability to reach the new state-of-the-art performance.

\section{Conclusion}

In this work, we introduce ChatKBQA, a generate-then-retrieve KBQA framework that utilizes advanced fine-tuned LLMs, which overcomes traditional challenges like retrieval inefficiencies, semantic parsing errors, and complexity of KBQA methods. Experimental results on WebQSP and CWQ benchmarks show that ChatKBQA achieves a new state-of-the-art KBQA performance. Its simplicity, flexibility, and plug-and-play make it an effective approach for combining LLM and KG in interpretable knowledge-required KBQA tasks.

\section*{Acknowledgments}
This work is supported by the National Science Foundation of China (Grant No. 62176026). This work is also supported by the BUPT Excellent Ph.D. Students Foundation (No. CX2023133) and the BUPT Postgraduate Innovation and Entrepreneurship Project (No. 2024-YC-A091).

\section*{Limitations}
% ChatKBQA is the first method to use fine-tuned LLMs to generate logical forms and then convert them into graph queries for KBQA tasks. Introducing the generate-then-retrieve approach solves many previous challenges and achieves state-of-the-art results. 

In Appendix~\ref{error}, we provide an error analysis of ChatKBQA, revealing significant room for improvement. Furthermore, we also discuss more limitations of ChatKBQA for future directions, which are detailed in Appendix~\ref{future}. 

\section*{Ethics Statement}
This paper investigates the problem of Knowledge Base Question Answering. We use large language models and retrieval methods to promote generation and retrieval performance. Therefore, we believe it does not violate any ethics.

\bibliography{custom}

\appendix

\clearpage
\appendix

\section*{Appendix} 

\section{Operators in Logical Form}
\label{operator}
Various operators include "AND" \texttt{(AND $E_1$ $E_2$)} to denote taking the intersection of $E_1$ and $E_2$, "COUNT" \texttt{(COUNT $E_1$)} to denote counting $E_1$, "ARGMAX" \texttt{(ARGMAX $E_1$ $r$)} to denote taking the max literal obtained after the projection of $E_1$ in the $r$ relation, "ARGMIN" \texttt{(ARGMIN $E_1$ $r$) } to denote taking the min literal obtained after the projection of the $r$ relation for $E_1$, "GT" \texttt{(GT $E_1$ $l$)} means to take the portion of $E_1$ that is greater than $l$, "GE" \texttt {(GE $E_1$ $l$)} to denote taking the part of $E_1$ greater than or equal to $l$, "LT" \texttt{(LT $E_1$ $l$)} to denote taking the part of $E_1$ less than $l$, "LE" \texttt{(LE $E_1$ $l$)} to denote taking the part of $E_1$ which is less than or equal to $l$, where $E_1$ or $E_2$ denote a sublayer logical form.

\section{Retrieval Complexity Analysis}
\label{complex}
During the retrieval phase, we measure the complexity of the algorithm using two indicators: the number of times vector similarity is calculated and the number of attempts to execute the logical form. Assuming the beam size in the generation phase is set to $b$, the size of the KB entity set is $E$, and the average logical form skeleton has $n_e$ entities, the complexity of entity retrieval is O($bn_e$$E$). For each entity's position, we select entities that rank in the top $k_e$ in similarity and are greater than the threshold $t_e$ for replacement. For the logical form as a whole, we select the top $k_1$ logical forms with a combined probability greater than the threshold $t_1$ as the result of entity retrieval.

In the relation retrieval phase, similarly, assuming the size of the KB relation set is R, and the average logical form skeleton has $n_r$ entities, the complexity of entity retrieval is O($k_1$$n_r$$R$). For each position's relation, we select relations that rank in the top $k_r$ in similarity and are greater than the threshold $t_r$ for replacement. For the logical form as a whole, based on the combination probability of the relation retrieval results, we select the top $k_2$ logical forms greater than the threshold $t_2$ as the result of relation retrieval.

Therefore, the complexity of the number of vector similarity calculations is O($b$$n_e$$E$ + $k_1$$n_r$$R$). For the number of attempts to execute the logical form, we initially attempt with the first b logical forms; if none can be executed, we proceed with entity retrieval and attempt up to $k_1$ times. If there is still no executable logical form, we move to relation retrieval and attempt up to $k_2$ times. Thus, the complexity of the number of logical form execution attempts is O($b$ + $k_1$ + $k_2$).

In this way, for KBQA tasks with large entity and relation sets, other parameters are much smaller than $E$ and $R$, making the complexity of vector similarity calculations in the order of O(n) and the complexity of logical form execution attempts in the order of O(1), both of which are controllable.

\section{Dataset Statistics}
\label{dataset}
As shown in Table~\ref{data}, this is the statistical information of the two KBQA datasets, WebQSP and CWQ, made by the ChatKBQA experiment.

\begin{table*}[h!t]
\small
  \centering
  \setlength{\tabcolsep}{2mm}{
    \begin{tabular}{rrrrrrrrr}
    \toprule
    \multicolumn{1}{c}{\textbf{Dataset}} & \multicolumn{1}{c}{\textbf{\#Question}} & \multicolumn{1}{c}{\textbf{\#Skeleton(LF)}} & \multicolumn{1}{c}{\textbf{\#Entity}} & \multicolumn{1}{c}{\textbf{\#Relation}} & \multicolumn{1}{c}{\textbf{\#Train}} & \multicolumn{1}{c}{\textbf{\#Valid}} & \multicolumn{1}{c}{\textbf{\#Test}} &
    \multicolumn{1}{c}{\textbf{KB}} \\
    \midrule
    \multicolumn{1}{c}{WebQSP} & \multicolumn{1}{c}{4,737} & \multicolumn{1}{c}{34} & \multicolumn{1}{c}{2,461} & \multicolumn{1}{c}{628} & \multicolumn{1}{c}{3,098} & \multicolumn{1}{c}{-} & \multicolumn{1}{c}{1,639} & 
    \multicolumn{1}{c}{Freebase} \\
    \multicolumn{1}{c}{CWQ} & \multicolumn{1}{c}{34,689} & \multicolumn{1}{c}{174} & \multicolumn{1}{c}{11,422} & \multicolumn{1}{c}{845} & \multicolumn{1}{c}{27,639} & \multicolumn{1}{c}{3,519} & \multicolumn{1}{c}{3,531} & 
    \multicolumn{1}{c}{Freebase} \\
    \bottomrule
    \end{tabular}}
\caption{\label{data}
Dataset statistics, where the columns respectively indicate the number of all KBQA questions, logical form skeletons, participant entities, participant relations, and questions in train/valid/test sets, followed by KB's name.}
\end{table*}%

\textbf{WebQSP dataset}~\citep{STAGG} is developed to evaluate the importance of gathering semantic parses compared to just answers for a set of questions. WebQSP consists of 4,737 KBQA questions, with 34 logical form skeletons and 2,461 entities involved. There are 628 relations specified within the dataset, which is divided into a training set of 3,098 questions and a test set of 1,639 questions. This dataset utilizes Freebase as its knowledge base and is tailored for developing systems that can process and answer natural language questions using structured data.

\textbf{CWQ dataset}~\citep{CWQ} is designed to answer complex questions requiring reasoning over multiple web snippets, which contains a large set of complex questions in natural language and is versatile in its applications. CWQ is considerably larger with 34,689 questions, underpinned by 174 logical form skeletons. It encompasses a more extensive set of entities amounting to 11,422 and includes 845 relations. The training set comprises 27,639 questions, supplemented by a validation set of 3,519 questions and a test set of 3,531 questions. CWQ also leverages Freebase as its knowledge base and is designed for complex question-answering tasks that require the interpretation and synthesis of information from various sources.

\section{More Baseline KBQA Methods}
\label{baseline}
% Appendix~\ref{hyper} shows the optimal hyperparameter settings for ChatKBQA.  Appendix~\ref{train} shows training details.
\begin{table}[h!t]
% \scriptsize
% \footnotesize
\small
\setlength{\tabcolsep}{0.6mm}{
\begin{tabular}{lcccccc}
\toprule
                                 & \multicolumn{3}{c}{\textbf{WebQSP}}                                                               & \multicolumn{3}{c}{\textbf{CWQ}}                                             \\ \cline{2-7} 
\multirow{-2}{*}{\textbf{Model}} & \textbf{F1}   & \textbf{Hits@1}  & \multicolumn{1}{c|}{\textbf{Acc}}  & \textbf{F1}   & \textbf{Hits@1} & \textbf{Acc}  \\ \hline
KV-Mem                           & 34.5          & 46.7                                         & \multicolumn{1}{c|}{-}             & 15.7          & 21.1                                         & -             \\
STAGG                           & 71.7          & -                                            & \multicolumn{1}{c|}{63.9}          & -             & -                                            & -             \\
GRAFT-Net                        & 62.8          & 67.8                                         & \multicolumn{1}{c|}{-}             & 32.7          & 36.8                                         & -             \\
UHop                             & 68.5          & -                                            & \multicolumn{1}{c|}{-}             & 29.8          & -                                            & -             \\
Topic Units                     & 67.9          & 68.2                                         & \multicolumn{1}{c|}{-}             & 36.5          & 39.3                                         & -             \\
TextRay                         & 60.3          & 72.2                                         & \multicolumn{1}{c|}{-}             & 33.9          & 40.8                                         & -             \\
PullNet                          & -             & 68.1                                         & \multicolumn{1}{c|}{-}             & -             & 47.2                                         & -             \\
QGG                             & 74.0          & 73.0                                         & \multicolumn{1}{c|}{-}             & 40.4          & 44.1                                         & -             \\
EmbedKGQA*                       & -             & 66.6                                         & \multicolumn{1}{c|}{-}             & -             & 44.7                                         & -             \\
EmQL*                            & -             & 75.5                                         & \multicolumn{1}{c|}{-}             & -             & -                                            & -             \\
NSM+h*                          & 67.4          & 74.3                                         & \multicolumn{1}{c|}{-}             & 44.0          & 48.8                                         & -             \\
GrailQA Ranking*                & 70.0          & -                                            & \multicolumn{1}{c|}{-}             & -             & -                                            & -             \\
ReTraCk*                         & 74.7          & 74.6                                         & \multicolumn{1}{c|}{-}             & -             & -                                            & -             \\
TransferNet                     & -             & 71.4                                         & \multicolumn{1}{c|}{-}             & -             & 48.6                                         & -             \\
Relation Learning               & 64.5          & 72.9                                         & \multicolumn{1}{c|}{-}             & -             & -                                            & -             \\
Rigel*                           & -             & 73.3                                         & \multicolumn{1}{c|}{-}             & -             & 48.7                                         & -             \\
CBR-KBQA                         & 72.8          & -                                            & \multicolumn{1}{c|}{69.9}          & 70.0          & 70.4                                         & 67.1          \\
Subgraph Retrieval*              & 64.1          & 69.5                                         & \multicolumn{1}{c|}{-}             & 47.1          & 50.2                                         & -             \\
RnG-KBQA                         & 75.6          & -                                            & \multicolumn{1}{c|}{71.1}          & -             & -                                            & -             \\
Program Transfer*                & 76.5          & 74.6                                         & \multicolumn{1}{c|}{-}             & 58.7          & 58.1                                         & -             \\
TIARA*                           & 78.9          & 75.2                                         & \multicolumn{1}{c|}{-}             & -             & -                                            & -             \\
UniK-QA                          & 79.1          & -                                            & \multicolumn{1}{c|}{-}             & -             & -                                            & -             \\
ArcaneQA                         & 75.6          & -                                            & \multicolumn{1}{c|}{-}             & -             & -                                            & -             \\
GMT-KBQA                         & 76.6          & -                                            & \multicolumn{1}{c|}{73.1}          & 77.0          & -                                            & 72.2          \\
Uni-Parser*                      & 75.8          & -                                            & \multicolumn{1}{c|}{71.4}          & -             & -                                            & -             \\
UnifiedSKG                       & 73.9          & -                                            & \multicolumn{1}{c|}{-}             & 68.8          & -                                            & -             \\
UniKGQA*                         & 72.2          & 77.2                                         & \multicolumn{1}{c|}{-}             & 49.4          & 51.2                                         & -             \\
DecAF                            & 78.8          & 82.1                                         & \multicolumn{1}{c|}{-}             & -             & 70.4                                         & -             \\
BeamQA*                          & -             & 73.4                                         & \multicolumn{1}{c|}{-}             & -             & -                                            & -             \\
HGNet*                           & 76.6          & 76.9                                         & \multicolumn{1}{c|}{70.7}          & 68.5          & 68.9                                         & 57.8          \\
SKP                              & -             & 79.6                                         & \multicolumn{1}{c|}{-}             & -             & -                                            & -             \\
StructGPT*                       & 72.6          & -                                            & \multicolumn{1}{c|}{-}             & -             & -                                            & -             \\
FC-KBQA                          & 76.9          & -                                            & \multicolumn{1}{c|}{-}             & 56.4          & -                                            & -             \\
Pangu                            & 79.6          & -                                            & \multicolumn{1}{c|}{-}             & -             & -                                            & -             \\ 
ToG*                         & -          & 82.6                                            & \multicolumn{1}{c|}{-}             & -         & 69.5                                            & -             \\ \hline\hline
ChatKBQA (ours)                        & 79.8          & 83.2                                         & \multicolumn{1}{c|}{73.8}          & 77.8          & 82.7                                         & 73.3          \\
ChatKBQA* (ours)                      & \textbf{83.5} & \textbf{86.4}                                & \multicolumn{1}{c|}{\textbf{77.8}} & \textbf{81.3} & \textbf{86.0}                                & \textbf{76.8} \\ \bottomrule
\end{tabular}}
\caption{\label{t4}
KBQA comparison of ChatKBQA with other baselines on WebQSP and CWQ datasets. * denotes using Oracle entity linking annotations. The results of the models are mainly taken from their original paper. For our proposed ChatKBQA framework, we display the results of the best setup on WebQSP and CWQ, respectively. The best results in each metric are in \textbf{bold}. 
}
\end{table}

% Existing Knowledge Base Question Answering (KBQA) methods can be broadly categorized into Information Retrieval-based (IR-based) and Semantic Parsing-based (SP-based) methods. (1) IR-based methods~\citep{KV-Mem,GRAFT-Net,PullNet,EmbedKGQA,NSM,TransferNet,RelationLearning,SubgraphRetrieval,UniK-QA,SKP} primarily retrieve relevant factual triples or text from Knowledge Bases (KBs) based on natural language questions, forming a subgraph to determine answers.
% (2) On the other hand, SP-based methods focus on translating questions into logical forms executable against KBs, such as SPARQL, query graph, and S-expression. Some SP-based approaches~\citep{STAGG,UHop,TopicUnits,TextRay,QGG,EMQL,GrailQA,Rigel,UniKGQA,BeamQA,HGNet,StructGPT,Pangu,ToG} utilize strategies of step-wise query graph generation and search for semantic parsing. Alternatively, other SP-based methods~\citep{ReTraCk,CBR-KBQA,RnG-KBQA,ProgramTransfer,TIARA,ArcaneQA,GMT-KBQA,Uni-Parser,UnifiedSKG,DECAF,FC-KBQA} employ sequence-to-sequence models to generate S-expressions completely and offer various enhancements to the semantic parsing process. In this paper, our proposed ChatKBQA is the first SP-based KBQA method using fine-tuned LLMs, which innovatively proposes a generate-then-retrieve approach to simplify KBQA method.

% In the main experiment, we compared ChatKBQA with all KBQA models in Section~\ref{s2} as follows in order of publication.

We compared ChatKBQA with more KBQA models as follows in order of publication.

\textbf{KV-Mem}~\citep{KV-Mem} uses a key-value structured memory model to enhance document comprehension and question-answering by encoding facts and reasoning over them for accurate predictions.

\textbf{STAGG}~\citep{STAGG} presents a KBQA method using semantic parse labeling, showing improvements in query accuracy compared to relying solely on question-answer pairs.

\textbf{GRAFT-Net}~\citep{GRAFT-Net} introduces a novel graph convolution-based neural network that enhances open-domain question answering by combining information from knowledge bases and text documents into a single model.

\textbf{UHop}~\citep{UHop} introduces a framework for unrestricted-hop relation extraction to handle queries requiring any number of relational hops in a knowledge graph, improving the capability to answer complex and indirect questions.

\textbf{Topic Units}~\citep{TopicUnits} utilizes a wide range of knowledge base units for question answering, employing a generation-and-scoring approach and reinforcement learning to enhance the identification and ranking of relevant topic units.

\textbf{TextRay}~\citep{TextRay} decomposes complex questions into simpler queries, processes them individually, and combines the results, using a semantic matching model.

\textbf{PullNet}~\citep{PullNet} presents a method that iteratively constructs a question-specific subgraph from knowledge bases and text for effective multi-hop reasoning in open-domain question answering.

\textbf{QGG}~\citep{QGG} introduces a method that enhances complex question answering by generating flexible query graphs for multi-hop questions and integrating constraints early.

\textbf{EmbedKGQA}~\citep{EmbedKGQA} introduces a method that uses knowledge graph embeddings to improve multi-hop question answering, addressing knowledge graph sparsity.

\textbf{EMQL}~\citep{EMQL} presents a method that combines centroid-sketch entity set representations with neural retrieval over embedded knowledge base triples.

\textbf{NSM$_{+h}$}~\citep{NSM} introduces a teacher-student framework for multi-hop KBQA, where the teacher network learns intermediate supervision signals through forward and backward reasoning to enhance the student network's reasoning capability.

\textbf{GrailQA Ranking}~\citep{GrailQA} presents a BERT-based KBQA model, demonstrating the critical role of pre-trained contextual embeddings, focusing on three levels of generalization - i.i.d., compositional, and zero-shot. 

\textbf{ReTraCk}~\citep{ReTraCk} introduces a neural semantic parsing framework, which combines retriever, transducer, and checker components for efficient and effective KBQA.

\textbf{TransferNet}~\citep{TransferNet} introduces a model that combines a transparent, attention-based approach with the ability to handle both label and text relations in a unified framework.

\textbf{Relation Learning}~\citep{RelationLearning} presents a method that integrates pre-trained language models with auxiliary tasks like relation extraction and reasoning.

\textbf{Rigel}~\citep{Rigel} introduces a method for enhancing end-to-end question answering using differentiable knowledge graphs, and adds an intersection operation to handle multiple-entity questions more effectively.

\textbf{CBR-KBQA}~\citep{CBR-KBQA} employs a case-based reasoning framework that retrieves similar cases (questions and logical forms) from a nonparametric memory, then reuses and revises these cases to generate logical forms for new questions, demonstrating its capability to handle complex questions and unseen relations without retraining.

\textbf{Subgraph Retrieval}~\citep{SR} introduces a method devising a trainable subgraph retriever (SR) decoupled from the reasoning process, which efficiently retrieves relevant subgraphs for question answering, enhancing performance by focusing on more relevant and smaller subgraphs and combining with subgraph-oriented reasoners.

\textbf{RnG-KBQA}~\citep{RnG-KBQA} introduces a framework that combines ranking and generation, using a rank-and-generate approach, where a ranker model identifies candidate logical forms and a generation model refines them.

\textbf{Program Transfer}~\citep{ProgramTransfer} proposes a novel two-stage parsing framework with an efficient ontology-guided pruning strategy for complex KBQA, which involves a sketch parser that translates questions into high-level program sketches and an argument parser that fills in detailed arguments.

\textbf{TIARA}~\citep{TIARA} introduces a novel method that enhances question answering over knowledge bases by using multi-grained retrieval, which improves the performance of pre-trained language models by focusing on the most relevant knowledge base contexts, including entities, logical forms, and schema items, and employs constrained decoding to control the output space, reducing generation errors and enhancing robustness in various generalization settings.

\textbf{UniK-QA}~\citep{UniK-QA} proposes a framework that integrates structured, unstructured, and semi-structured knowledge sources, such as text, tables, lists, and knowledge bases, which flattens all data into text and applies a unified retriever-reader model.

\textbf{ArcaneQA}~\citep{ArcaneQA} introduces a generation-based KBQA model that addresses large search space and schema linking challenges in KBQA, which employs dynamic program induction for efficient search space navigation and dynamic contextualized encoding for improved schema linking.

\textbf{GMT-KBQA}~\citep{GMT-KBQA} proposes a multi-task learning framework with a shared T5 encoder to improve question answering over knowledge bases by simultaneously learning entity disambiguation, relation classification, and logical form generation.

\textbf{Uni-Parser}~\citep{Uni-Parser} unifies semantic parsing for question answering on both knowledge bases and databases by using a three-module approach: primitive enumeration, ranking, and compositional generation.

\textbf{UnifiedSKG}~\citep{UnifiedSKG} unifies 21 structured knowledge grounding tasks into a text-to-text format, leveraging T5 models and multi-task learning to improve performance across diverse tasks and facilitate zero-shot and few-shot learning investigations.

\textbf{UniKGQA}~\citep{UniKGQA} integrates retrieval and reasoning for multi-hop question answering over knowledge graphs, employing a unified architecture that combines a semantic matching module and a matching information propagation module, enhanced by pre-training and fine-tuning strategies.

\textbf{DecAF}~\citep{DECAF} combines the generation of logical forms and direct answers, leveraging a sequence-to-sequence framework with retrieval from linearized knowledge bases.

\textbf{BeamQA}~\citep{BeamQA} combines sequence-to-sequence prediction and beam search for multi-hop knowledge graph question answering, using a fine-tuned BART model for path generation and a novel beam search execution algorithm to traverse the knowledge graph and find answers.

\textbf{HGNet}~\citep{HGNet} proposes a hierarchical query graph generation approach with an outlining stage for structural constraints and a filling stage for instance selection.

\textbf{SKP}~\citep{SKP} introduces structured knowledge-aware pre-training tasks, an efficient linearization strategy, and an interval attention mechanism, leading to significant improvements in subgraph retrieval and encoding.

\textbf{StructGPT}~\citep{StructGPT} enhances LLMs' reasoning over structured data using an Iterative Reading-then-Reasoning (IRR) approach, which includes specialized interfaces for efficient data access, a novel invoking-linearization-generation procedure, and iterative reasoning to effectively utilize structured data in answering complex questions.

\textbf{FC-KBQA}~\citep{FC-KBQA} introduces a Fine-to-Coarse composition framework for question answering over knowledge bases, utilizing fine-grained component detection, middle-grained component constraints, and coarse-grained component composition.

\textbf{Pangu}~\citep{PanGu} proposes a grounded language understanding framework that combines a symbolic agent and a neural language model, which allows for the incremental construction of valid plans and utilizes the language model to evaluate the plausibility of these plans.

\textbf{ToG}~\citep{ToG} integrates LLMs with KGs for deep and responsible reasoning, using a beam search algorithm in KG/LLM reasoning, which allows the LLM to dynamically explore multiple reasoning paths in KG and make decisions accordingly, enhancing LLMs' deep reasoning capabilities for knowledge-intensive tasks.

\section{Hyperparameter Settings}
\label{hyper}

We use the grid search method to select the optimal hyperparameter settings for the network. The F1 score of KBQA predicted without oracle entity linking is chosen as the evaluation metric. The hyperparameters that we can adjust and the possible values of the hyperparameters are first determined according to the structure of our model in Table~\ref{t6}. 

Afterward, the different hyperparameter choices are combined to judge the merit of the hyperparameter combinations. The optimal hyperparameter combinations of the model are obtained by circular traversal of all combinations. The optimal hyperparameter combinations are shown in \textbf{bold}.

\begin{table*}[h!t]
%\small
%\footnotesize
% \scriptsize
\small
\centering
\begin{tabular}{rrrr}
\toprule
\multicolumn{1}{c}{\textbf{Hyperparameter}} & \multicolumn{1}{c}{\textbf{WebQSP}}& \multicolumn{1}{c}{\textbf{CWQ}}\\
\midrule
\multicolumn{1}{c}{LLM Selection} & \multicolumn{1}{c}{\textbf{Llama-2-7B}}& \multicolumn{1}{c}{\textbf{Llama-2-13B}}\\
\multicolumn{1}{c}{Fine-tuning Type} & \multicolumn{1}{c}{$\left\{\textbf{\text{LoRA}}, \text{QLoRA}, \text{P-tuning v2}, \text{Freeze}\right\}$}& \multicolumn{1}{c}{$\left\{\textbf{\text{LoRA}}, \text{QLoRA}, \text{P-tuning v2}, \text{Freeze}\right\}$}\\
\multicolumn{1}{c}{Train Batch Size} & \multicolumn{1}{c}{$\left\{1, 2, 3, \textbf{4}\right\}$}& \multicolumn{1}{c}{$\left\{1, 2, 3, \textbf{4}\right\}$}\\
\multicolumn{1}{c}{Learning Rate} & \multicolumn{1}{c}{$\left\{\textbf{\text{5e-5}}, \text{5e-4}, \text{5e-3}\right\}$}& \multicolumn{1}{c}{$\left\{\textbf{\text{5e-5}}, \text{5e-4}, \text{5e-3}\right\}$}\\
\multicolumn{1}{c}{Train Epoch} & \multicolumn{1}{c}{$\left\{10, 50, \textbf{100}\right\}$}& \multicolumn{1}{c}{$\left\{\textbf{10}, 50, 100\right\}$}\\
\multicolumn{1}{c}{Test Batch Size} & \multicolumn{1}{c}{$\left\{\textbf{1}, 2, 3, 4\right\}$}& \multicolumn{1}{c}{$\left\{\textbf{1}, 2, 3, 4\right\}$}\\
\multicolumn{1}{c}{Beam Size} & \multicolumn{1}{c}{$\left\{1, 2, 5, 8, \textbf{15}\right\}$}& \multicolumn{1}{c}{$\left\{1, 2, 5, \textbf{8}\right\}$}\\
\multicolumn{1}{c}{Retrieval Type} & \multicolumn{1}{c}{$\left\{\textbf{\text{SimCSE}}, \text{Contriever}, \text{BM25}\right\}$}& \multicolumn{1}{c}{$\left\{\textbf{\text{SimCSE}}, \text{Contriever}, \text{BM25}\right\}$}\\
\multicolumn{1}{c}{ER Top $k_e$} & \multicolumn{1}{c}{$\left\{5, 10, \textbf{50}, 100\right\}$}& \multicolumn{1}{c}{$\left\{5, 10, \textbf{50}, 100\right\}$}\\
\multicolumn{1}{c}{ER Threshold $t_e$} & \multicolumn{1}{c}{$\left\{0.0, 0.0001, \textbf{0.001}, 0.01, 0.1\right\}$}& \multicolumn{1}{c}{$\left\{0.0, 0.0001, \textbf{0.001}, 0.01, 0.1\right\}$}\\
\multicolumn{1}{c}{ER Top $k_1$} & \multicolumn{1}{c}{$\left\{10, 30, \textbf{50}, 100, 1000\right\}$}& \multicolumn{1}{c}{$\left\{10, \textbf{30}, 50, 100, 1000\right\}$}\\
\multicolumn{1}{c}{ER Threshold $t_1$} & \multicolumn{1}{c}{$\left\{\textbf{0.0}, 0.0001, 0.001, 0.01, 0.1\right\}$}& \multicolumn{1}{c}{$\left\{\textbf{0.0}, 0.0001, 0.001, 0.01, 0.1\right\}$}\\
\multicolumn{1}{c}{RR Top $k_r$} & \multicolumn{1}{c}{$\left\{3, 5, \textbf{15}, 30\right\}$}& \multicolumn{1}{c}{$\left\{3, 5, \textbf{15}, 30\right\}$}\\
\multicolumn{1}{c}{RR Threshold $t_r$} & \multicolumn{1}{c}{$\left\{0.0, 0.0001, 0.001, \textbf{0.01}, 0.1\right\}$}& \multicolumn{1}{c}{$\left\{0.0, 0.0001, 0.001, \textbf{0.01}, 0.1\right\}$}\\
\multicolumn{1}{c}{RR Top $k_2$} & \multicolumn{1}{c}{$\left\{30, \textbf{300}, 3000, 10000\right\}$}& \multicolumn{1}{c}{$\left\{40, 400, \textbf{4000}, 10000\right\}$}\\
\multicolumn{1}{c}{RR Threshold $k_2$} & \multicolumn{1}{c}{$\left\{\textbf{0.0}, 0.0001, 0.001, 0.01, 0.1\right\}$}& \multicolumn{1}{c}{$\left\{\textbf{0.0}, 0.0001, 0.001, 0.01, 0.1\right\}$}\\
\bottomrule   
\end{tabular}

\caption{\label{t6}
Hyperparameter Search.
}

\end{table*}

For example, WebQSP hyperparameter choices select the Llama-2-7B model, as shown by bolded values, for optimal model performance. LoRA is the fine-tuning type chosen, suggesting low-rank adjustments to model parameters. A train batch size of 4, learning rate of 5e-4, and 50 training epochs indicate a preference for moderate-sized data processing batches and a faster learning rate over many epochs. Test batch size of 4 and beam size of 5 indicate evaluation and prediction generation configuration. The retrieval algorithm was SimCSE because it compares sentence embeddings well. The top-k and threshold values for Entity Retrieval (ER) and Relation Retrieval (RR) were set to balance retrieving relevant information and computational efficiency.

\section{Effectiveness of Beam Search}
\label{beam}
Beam search is a heuristic algorithm usually used in sequence generation tasks, which expands the search space by generating multiple highly probable logical forms instead of only one. As shown in Figure~\ref{b}, an increase in beam size enhances the likelihood of executing SPARQL queries based on candidate logical forms, improving the KBQA performance.

\section{Plug-and-Play Settings}
\label{play}
ChatKBQA has a plug-and-play characteristic, as shown in 3 parts, including the Open-source LLMs, PEFT methods, and Unsupervised Retrieval methods, all of which have different candidates. The following is a description of these candidates.

\subsection{Open-source Large Language Models}
In the open-sourced macro modelling part, we choose Llama-2, ChatGLM2, and Baichuan2.

\textbf{Llama-2-7B / Llama-2-13B}~\citep{Llama2}: Part of Meta AI's Llama series, these models are auto-regressive transformers with 7 and 13 billion parameters, trained on 2 trillion tokens. They are optimized for dialogue and general language tasks, leveraging supervised fine-tuning and reinforcement learning for better alignment with human preferences.

\textbf{ChatGLM2-6B}~\citep{GLM}: Developed by Tsinghua University, this 6.2 billion-parameter bilingual Chinese-English chat model improves upon its predecessor with enhanced performance, longer context support, and efficient inference. It's designed for fluent, coherent conversations in both languages.

\textbf{Baichuan2-7B / Baichuan2-13B}~\citep{Baichuan2}: From Baichuan Intelligent Technology, these multilingual models have 7 and 13 billion parameters and are trained on 2.6 trillion tokens. They support Chinese and English, offering competitive performance on various language processing benchmarks and are available for open-source commercial use.

\subsection{Parameter-Efficient Fine-Tuning Methods}
In the PEFT part, we choose LoRA, QLoRA, P-tuning v2, and Freeze.

\textbf{LoRA (Low-Rank Adaptation)}~\citep{LoRA} is a PEFT method that introduces low-rank matrices to adapt large pre-trained models. Instead of fine-tuning all parameters, LoRA modifies only a small number of additional trainable parameters, effectively reducing the computational cost. It alters the weights of a pre-trained model in a low-rank decomposed space, allowing for efficient adaptation while maintaining the original model's structure and size.

\textbf{QLoRA (Quantized Low-Rank Adaptation)}~\citep{QLoRA} is an extension of LoRA, combining low-rank adaptation with quantization techniques. It aims to further reduce the computational and memory overhead associated with fine-tuning large models. By quantizing the additional low-rank matrices introduced in LoRA, QLoRA provides a more memory-efficient approach to adapting pre-trained models. 

\textbf{P-tuning v2}~\citep{P-Tuningv2} advances the concept of prompt tuning, where trainable prompts are added to a fixed pre-trained model to guide its predictions. P-tuning v2 introduces trainable continuous prompts at the embedding layer and employs a sophisticated bi-level optimization strategy. This approach enhances the model's ability to adapt to specific tasks with minimal parameter updates, making it more efficient than traditional fine-tuning methods.

\textbf{Freeze}~\citep{Freeze} is a parameter-efficient approach where most of the layers of a pre-trained model are frozen, and only a small fraction of the parameters are fine-tuned. This technique significantly reduces the computational resources required for fine-tuning, making it ideal for scenarios with limited budgets. By selectively updating only certain layers or parts of a model, Freeze retains the general knowledge of the pre-trained model while adapting it to specific tasks.

\subsection{Unsupervised Retrieval Methods}
In the Unsupervised Retrieval part, we choose SimCSE, Contriever and BM25.

\textbf{SimCSE}~\citep{SimCSE} is an unsupervised method for generating sentence embeddings using contrastive learning. It enhances semantic understanding by using variations of the same sentence to train neural networks, improving performance in tasks like textual similarity and natural language inference.

\textbf{Contriever}~\citep{Contriever} is an unsupervised technique for creating dense passage embeddings, designed for effective retrieval in large document collections. It focuses on semantic content, offering an advanced alternative to traditional keyword-based retrieval methods. 
 
\textbf{BM25}~\citep{BM25} is a probabilistic ranking function used in search engines. It evaluates document relevance to a search query, improving upon models like TF-IDF by incorporating document length normalization and term frequency saturation.

\section{Error Analysis}
\label{error}
We analyze the questions in the WebQSP test set that were not answered correctly by ChatKBQA without oracle entity linking, and errors can be summarized as follows.

\textbf{Logical form skeleton error }(40.10\%).
We discover that the majority of the errors are caused by ChatKBQA failing to provide the correct logical form skeleton for the question, e.g. predicting "\texttt{(JOIN (R []) (JOIN (R []) []))}" as "\texttt{(JOIN (R []) [])}". This is due to the limited representation of certain complex skeletons in train set. 

% It promises to investigate better approaches for creating training sets to improve semantic parsing.

\textbf{Entity retrieval error }(27.17\%).
Then, a portion of the samples that predicted the correct logical form skeletons, but did not retrieve the correct entities, e.g. predicting "\texttt{(JOIN (R []) m.0d3k14)}" as "\texttt{(JOIN (R []) m.07618sw)}". 

\textbf{Relation retrieval error }(19.48\%).
In the case of successful skeleton prediction and entity retrieval, errors in relation retrieval can also lead to failed logical form generation that does not match the ground truth, e.g. predicting "\texttt{(JOIN (R finance.currency.countries\_used) m.0kz1h)}" as "\texttt{(JOIN (R finance.currency. currency\_code) m.0kz1h)}". 

\textbf{SPARQL convertion error }(13.26\%).
Finally, a small proportion of the remaining errors arise from the fact that, although the generated logical form is consistent with the ground truth, it fails to execute or the answers are inconsistent when converted to SPARQL, which may be caused by the loss of the conversion from logical form to SPARQL.% or possibly changes in the KB.

\section{Discussion of LLM combined with KG. }
\label{future}
\subsection{Insights from ChatKBQA. }

(1) We propose a straightforward KBQA framework that uses fine-tuned open-source large models for the first time. (2) Innovatively, we adopt a generate-then-retrieve approach to enhance generation outcomes and retrieval efficiency separately, ultimately boosting KBQA performance. (3) Our framework has plug-and-play capabilities, allowing flexible replacement of LLMs and retrieval models to address the KBQA challenge. (4) Our approach introduces a new paradigm for LLMs to conduct interpretable knowledge-based Q\&A, offering a fresh perspective on merging LLMs and KGs.

To summarize, ChatKBQA proposes a thought taking both the advantages of using LLMs to do natural language semantic parsing for graph query generation and calling external KBs to interpretably reason with queries, which we name Graph Query of Thoughts (GQoT), a promising LLM+KG combination paradigm to better utilize the external knowledge, improve Q\&A's interpretability, and avoid LLM's hallucinations.

\subsection{Future Directions. }
ChatKBQA still has much room for improvement, such as in the design of the training set, the decomposition of complex questions, support for various graph query languages, and applications in specific domains, which are our future research directions:

\textbf{Training set design}: ChatKBQA is the first method to fine-tune open-source large models using unsupervised retrieval methods for the KBQA task, achieving state-of-the-art results. Therefore, the effectiveness of fine-tuning depends on the quality of the dataset used to map natural language to logical forms. In future work, we plan to enhance the training set by extracting computation graphs from the knowledge graph using graph sampling, then converting them into natural language, and exploring ways to achieve maximum training effectiveness with the least amount of training data.

\textbf{Decomposition of complex questions}: We have seen that for some simple tasks, such as one-hop and two-hop queries, ChatKBQA performs very well because the logical form skeletons involved are very similar and the fine-tuned LLM can generate them effectively. However, generating the corresponding long logical forms for more complex questions is a challenge. Therefore, in future work, we plan to use techniques such as CoT or Agent to decompose natural language questions into simpler logical forms for better performance.

\textbf{Support for various graph query languages}: Currently, ChatKBQA converts generated logical forms into SPARQL queries in two datasets, as the Freebase KB stores knowledge in RDF format. We will explore more KBs and datasets, such as those using the Cypher language like Neo4j, where the methodology of generating and then retrieving with ChatKBQA is also promising.

\textbf{Open-domain and specific-domain applications}: There is a demand for precision knowledge question answering in fields such as open-domain, medicine, finance, and telecommunications. We can first use UIE or LLM information extraction technology to build a knowledge graph, then fine-tune ChatKBQA to understand the structure of the knowledge graph, achieving interpretable knowledge Q\&A in open and specific domains.

\end{document}